\documentclass{article}

 \usepackage[dblblindworkshop, final]{neurips_2025}
\workshoptitle{Mechanistic Interpretability Workshop}

\usepackage[utf8]{inputenc} 
\usepackage[T1]{fontenc}    
\usepackage{hyperref}       
\usepackage{url}            
\usepackage{booktabs}       
\usepackage{amsfonts}       
\usepackage{nicefrac}       
\usepackage{microtype}      
\usepackage{xcolor}         
\usepackage{amsmath}
\usepackage{graphicx}
\title{Universal Neurons in GPT-2: Emergence, Persistence, and Functional Impact}

\author{%
  Advey Nandan\thanks{Equal Contribution} \\
  University of Waterloo\\
  \texttt{adveynandan@gmail.com} \\
  \And
  Cheng-Ting Chou\footnotemark[1] \\
  University of Illinois Urbana-Champaign \\
  \texttt{ctchou3@illinois.edu} \\
  \And
  Amrit Kurakula \\
  Ottawa University \\
  \texttt{amritlalith345@gmail.com} \\
  \And
  Cole Blondin \\
  Algoverse \\
  \texttt{cole@algoverseairesearch.org} \\
  \And
  Kevin Zhu \\
  Algoverse \\
  \texttt{kevin@algoverseacademy.com} \\
  \And
  Vasu Sharma \\
  Meta FAIR Lab \\
  \texttt{sharma.vasu55@gmail.com} \\
  \And
  Sean O’Brien \\
  Algoverse \\
  \texttt{seobrien@ucsd.edu} \\
}

\begin{document}

\maketitle

\begin{abstract}
We investigate the phenomenon of neuron universality in independently trained GPT-2 Small models, examining these universal neurons—neurons with consistently correlated activations across models—emerge and evolve throughout training. By analyzing five GPT-2 models at five checkpoints, we identify universal neurons through pairwise correlation analysis of activations over a dataset of 5 million tokens. Ablation experiments reveal significant functional impacts of universal neurons on model predictions, measured via cross entropy loss. Additionally, we quantify neuron persistence, demonstrating high stability of universal neurons across training checkpoints, particularly in early and deeper layers. These findings suggest stable and universal representational structures emerge during language model training.
\end{abstract}

\section{Introduction}

Large language models (LLMs) exhibit remarkable generalization but remain difficult to interpret \citep{Bommasani2021FoundationModels}. However, neural networks are fully observable and deterministic, allowing us to record and manipulate internal components such as neuron activations \citep{bricken2023monosemanticity}. This presents a rare opportunity to reverse-engineer their internal mechanism. An important open question regarding interpretability is whether models independently trained on the same task converge on similar internal structures—a notion termed the \textit{universality hypothesis} \citep{wang2024towards}. Universality, if established, offers stable interpretability targets and aids transfer learning.

We examine this hypothesis by analyzing five GPT-2 models trained from scratch, tracking when universal neurons—units with highly correlated activations across models \citep{gurnee2024universal}—emerge, their stability over training, and their causal role.

\textbf{Contributions:}
\begin{itemize}
\item \textbf{Emergence Analysis}: We provide the first systematic study on the emergence of universal neurons during training, showing that they appear early and grow steadily, especially in early and deeper layers.
\item \textbf{Persistence Quantification}: We quantify the stability of universal neurons across training checkpoints, finding that most remain universal in subsequent stages.
\item \textbf{Functional Role via Ablation}:We demonstrate that ablating universal neurons significantly increases loss, confirming their causal importance to model predictions.
\item \textbf{Layer-wise Characterization}: We show that first-layer universal neurons disproportionately affect output distributions, suggesting they encode critical low-level information.
\end{itemize}

\section{Related Work}

\paragraph{Universality and Cross-Model Consistency.}
Early studies reported limited direct neuron matching \citep{pmlr-v97-kornblith19a}. However, recent work identifies universal neurons with consistent semantic features across independently trained GPT-2 models \citep{gurnee2024universal}. These universal neurons are also shown to correspond to interpretable, semantically meaningful features \citep{gurnee2024universal}. This provides evidence that some circuits are consistently discovered across training runs, supporting the hypothesis of shared representational scaffolding.

\paragraph{Representation Similarity.}
Because neurons are often polysemantic \citep{bricken2023monosemanticity}, direct comparison is difficult. \citep{lan2025sparseautoencodersrevealuniversal} addressed this by learning sparse features with autoencoders and found significant alignment of feature dimensions across models. Algorithmic behaviors also show cross-architecture consistency, indicating broader universality \citep{lindsey2025biology, ameisen2025circuit}.

\paragraph{Emergence and Stability.}
Works such as the lottery ticket hypothesis \citep{frankle2018the} and canonical correlation studies \citep{NIPS2017_7188} show that networks form persistent representational patterns within the first few training epochs. Theoretical analyses further support the idea that networks rapidly learn dominant features that are gradually refined \citep{saxe2019mathematical}. These results motivate our focus on the \textit{emergence} and \textit{persistence} of universal neurons throughout training.

\section{Method}

We analyze five GPT-2 Small models at checkpoints 20\%, 40\%, 60\%, 80\%, and 100\% of training(80, 160k, 240k, 320k, 400k steps). Neuron activations are extracted over 5M tokens from the Pile dataset \citep{monology_pile_uncopyrighted}.

\paragraph{Identifying Universal Neurons via Correlation.}

Following \citep{gurnee2024universal}, we compute Pearson correlations between neurons across model pairs. Let $\mathbf{a}_k^{(m, c)} \in \mathbb{R}^n$ denote the activation vector of neuron $k$ in model $m$ at checkpoint $c$ over $n$ token positions. The Pearson correlation between neurons defined by $\mathbf{a}_k^{(m_1, c)}$ and $\mathbf{a}_\ell^{(m_2, c)}$ is:

\[
\rho_{k, \ell}^{(m_{1},m_{2}, c)} = \frac{\mathbb{E}\left[(\mathbf{a}_k^{(m_{1}, c)} - \mu_k)(\mathbf{a}_\ell^{(m_{2}, c)} - \mu_\ell)\right]}{\sigma_k \sigma_\ell}
\]

where $\mu_k$ and $\sigma_k$ are mean and standard deviations of the activation vector $\mathbf{a}_k^{(m, c)}$ computed across a 5 million token dataset of the uncopyrighted Pile HuggingFace dataset \citep{monology_pile_uncopyrighted}. We compute the excess correlation for a neuron $k$ with respect to a model $m_2$ at checkpoint $c$ as:

\[
\varrho_{k, m_2, c} = \left( 
\max_{\ell \in N(m_2)} \rho^{m_1,m_2, c}_{k,\ell} - 
\max_{\ell \in N_R(m_2)} \bar{\rho}^{m_1,m_2, c}_{k,\ell} 
\right)
\]

where $\bar{\rho}^{m_1,m_2,c}_{k,\ell} $ is the pearson correlation between  neuron $k$ in model $m_1$ and neuron $\ell$ in a randomly rotated version of the layer from model $m_2$, all at a checkpoint $c$. This rotation is constructed by multiplying the matrix of activations in that layer with a random Gaussian matrix, as described in \citep{gurnee2024universal}. The purpose of this transformation of activation vectors is to eliminate any privileged basis and establish a baseline for comparison \citep{gurnee2024universal}.

We used five GPT-2 Small models, models \texttt{a} through \texttt{e} \footnote{from stanford-crfm at HuggingFace}. We selected model \texttt{a} as the reference and computed Pearson correlations between its neurons and those in each of the other models (\texttt{b}, \texttt{c}, \texttt{d}, \texttt{e}). A neuron in model \texttt{a} is labeled \textit{universal} if it averages an excess correlation above 0.5 across all 4 model pairs. We also adjust this threshold to 0.4 and 0.6 to verify robustness. Tracking $\varrho_i$ across checkpoints and models allows us to observe universality emerge over training.

\paragraph{Persistence Across Training Checkpoints.}

We evaluate whether neurons remain universal over time by computing:

\[
P_{\text{persist}} = P(\text{univ. at } t_2 \mid \text{univ. at } t_1)
\]

across training step intervals (e.g., 80k$\rightarrow$160k, 240k$\rightarrow$320k). To localize this further, we stratify by transformer layer $\ell$:

\[
P_{\text{persist}}(\ell) = P(\text{univ.}_{t_2} \mid \text{univ.}_{t_1}, \text{layer} = \ell)
\]

\paragraph{Ablation Studies and Functional Role.}

To test functional significance, we ablate universal and control (non-universal) neurons during inference by zeroing their MLP outputs. We then measure changes in loss and change in loss per neuron ablated. We also perform a sensitivity analysis by repeating the experiment with relaxed thresholds for determining universality (e.g., 0.4 and 0.6).

\section{Results}

\begin{figure}[b]
    \centering
    \includegraphics[width=0.8\linewidth]{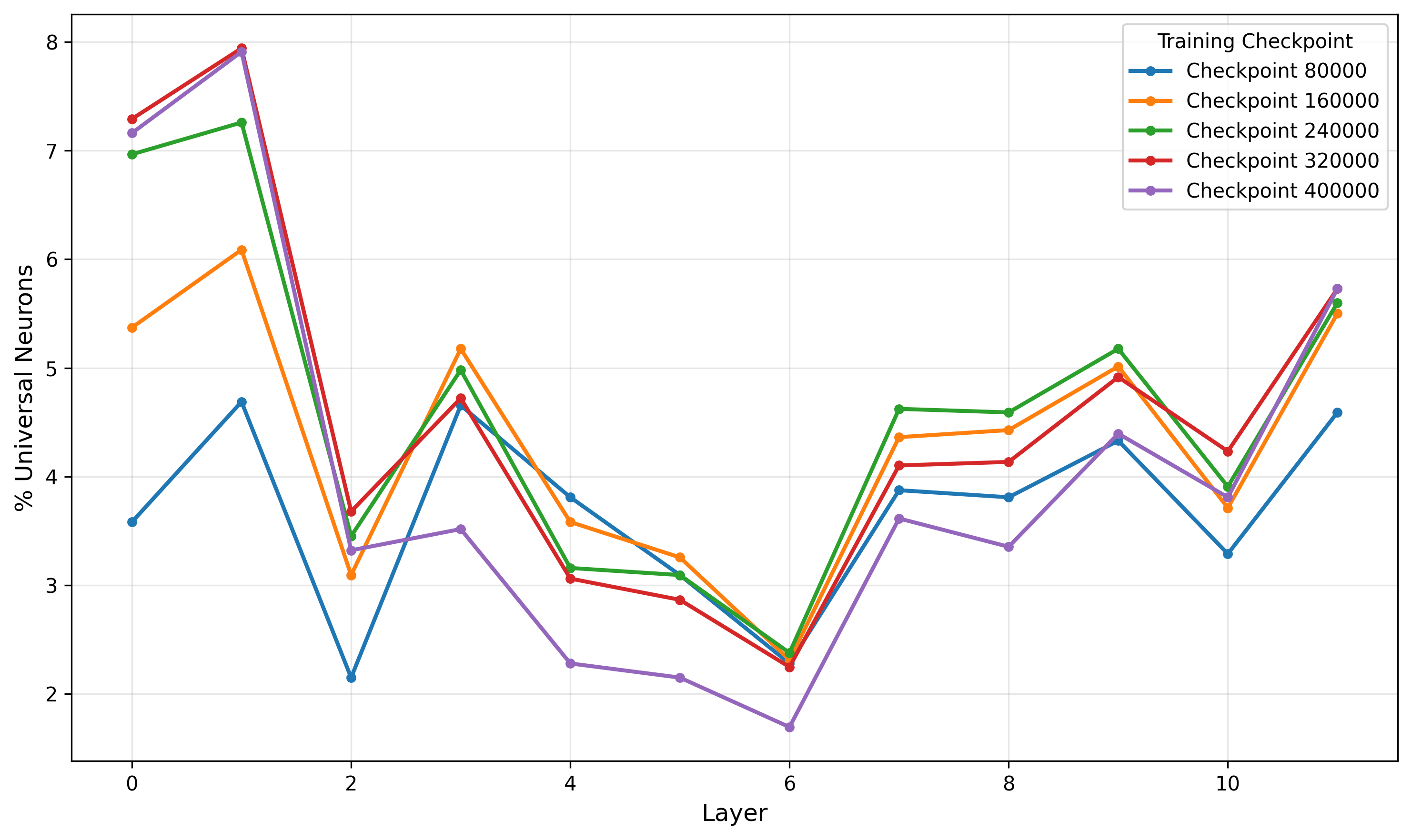}
    \caption{Percentage of Universal Neurons Across Layers. The graph shows an increasing trend of Universal Neurons as training step increases.}
    \label{fig:dis}
\end{figure}

\paragraph{Emergence of Universal Neurons.}
Universal neurons emerge early, increasing consistently through training, notably in earlier layers (Figure \ref{fig:dis}). At early checkpoints (80k steps), fewer than 5\% of neurons meet the universality criterion (0.5 threshold), but this fraction increases steadily by 400k steps. As shown in Appendix~\ref{app:appA}, adjusting the universality threshold to 0.4 or 0.6 shows consistent trends.

\paragraph{Persistence of Universal Neurons.}
We assess how universal neurons remain universal consistently as training progresses by computing their conditional persistence in five intervals: 80k→160k, 160k→240k, 240k→320k, 320k→400k, and 80k→400k steps.

\begin{figure}[t]
    \centering
    \includegraphics[width=0.8\linewidth]{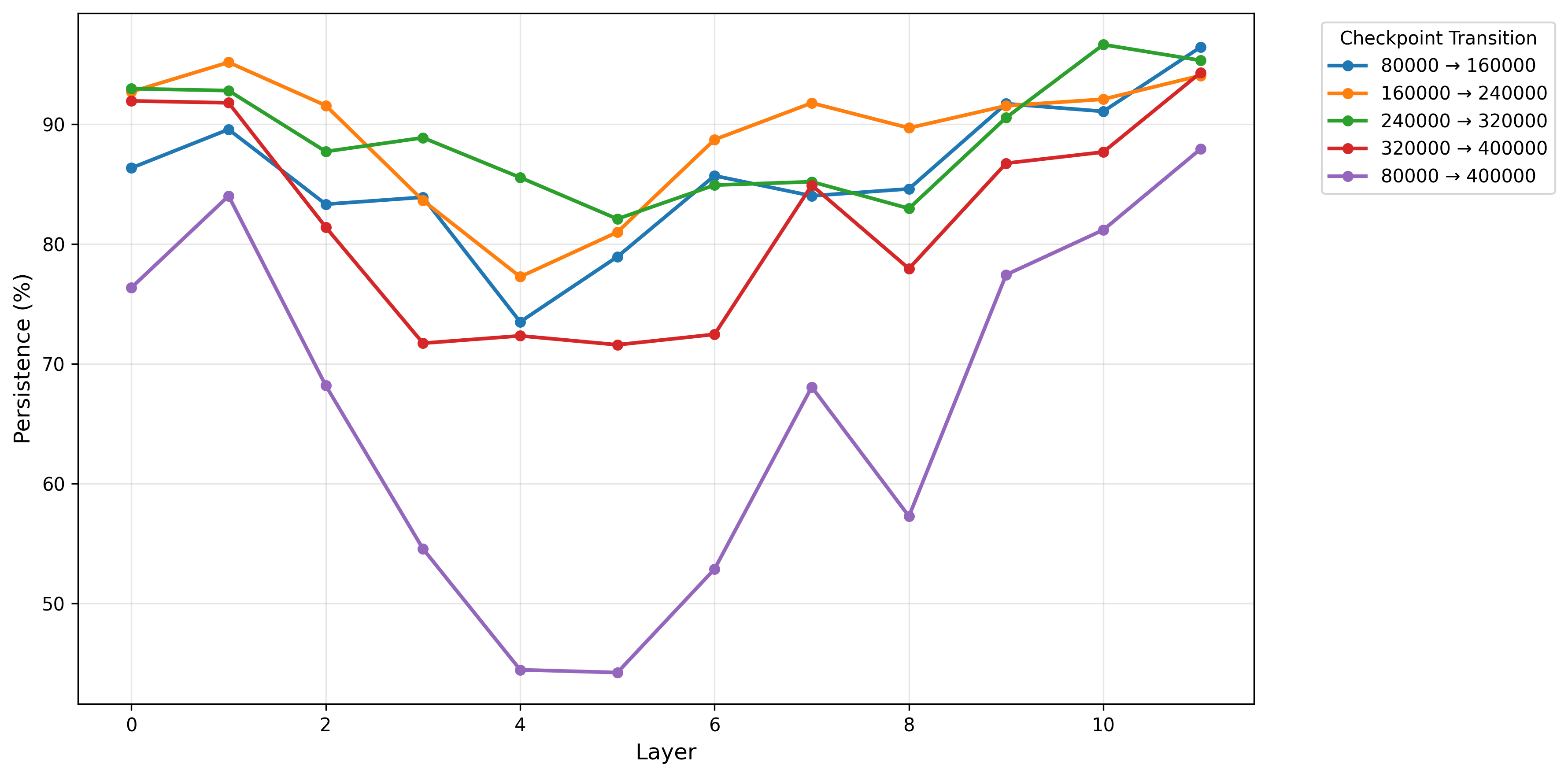}
    \caption{Universal Neuron Persistence Across Layers. Early and later layers show high Universal Neuron persistence, while middle layers experience shifting dynamics of universal features.}
    \label{fig:persist}
\end{figure}

Figure~\ref{fig:persist} shows layer-wise persistence rates across these intervals. We find that universal neurons are mostly highly stable over time, especially in early and later layers. Layers 0, 1, 10, and 11 maintain near or above 80\% persistence, while mid-layer neurons (e.g., layers 3–8) show greater volatility. The overall persistence of 80k→400k is in general lower than the adjacent intervals, reflecting a gradual representational drift.

These results support the hypothesis that universal neurons, particularly in early and deeper layers, encode stable and task-relevant features that solidify as training proceed. For completeness, we include layer-wise persistence plots in Appendix~\ref{app:appB} with different thresholds. 

\paragraph{Ablating Universal Neurons.}
\label{sec:result-ablate}
To test the functional importance of universal neurons, we ablate them by zeroing their MLP outputs during inference and measure the resulting change in model predictions using Cross Entropy Loss. We measure the loss value of ablating groups of neurons and the ablation efficiency: change in loss per neuron ablated.

\begin{figure}[b]
  \centering
  \begin{minipage}[t]{0.49\linewidth}
    \centering
    \includegraphics[width=\linewidth]{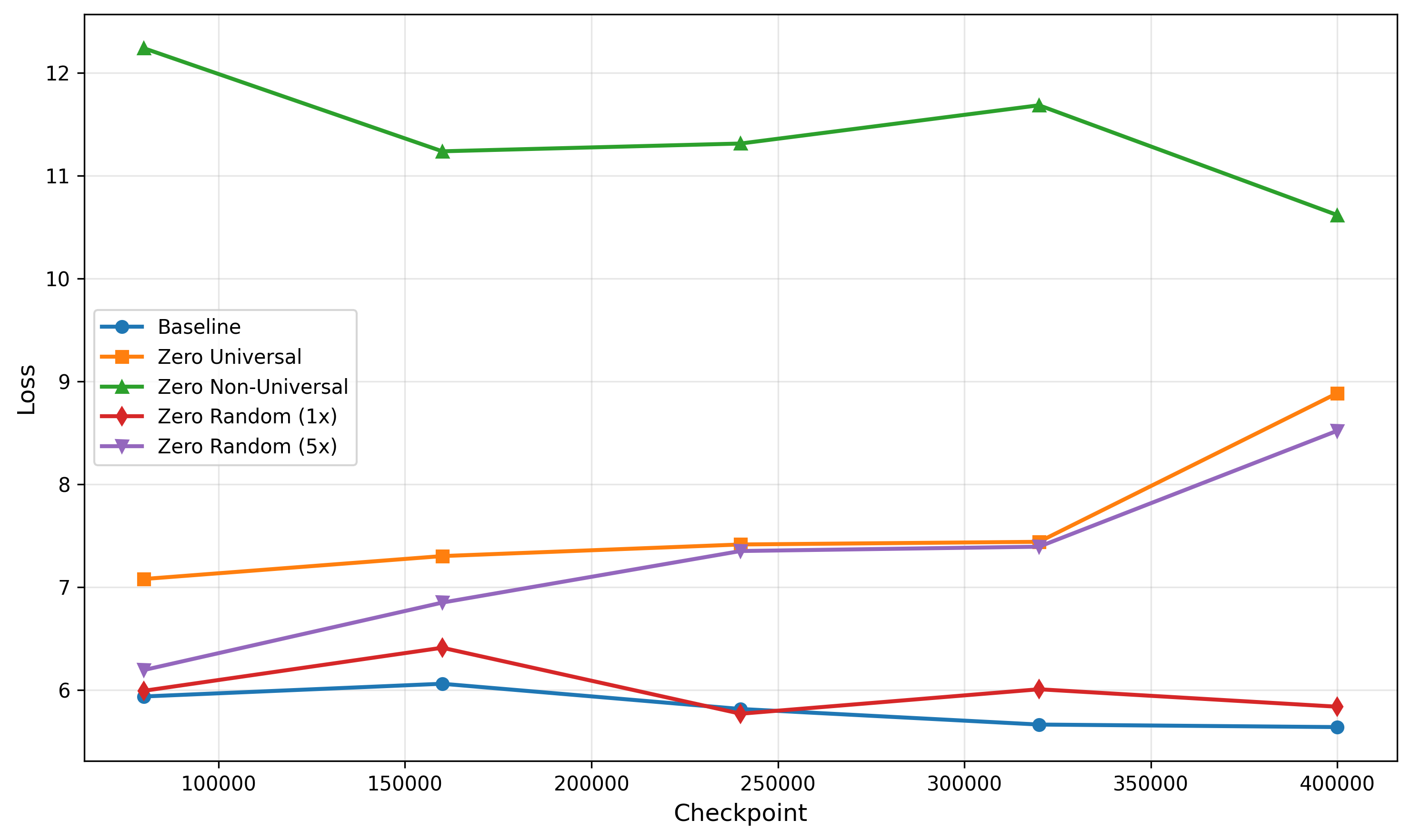}
    \caption{Absolute Loss Values After Ablating(zeroing activations) Different Neurons. It takes around 5x the amount of random neurons to achieve the same disruptive result of ablating universal neurons.}
    \label{fig:ablate_loss}
  \end{minipage}%
  \hfill
  \begin{minipage}[t]{0.49\linewidth}
    \centering
    \includegraphics[width=\linewidth]{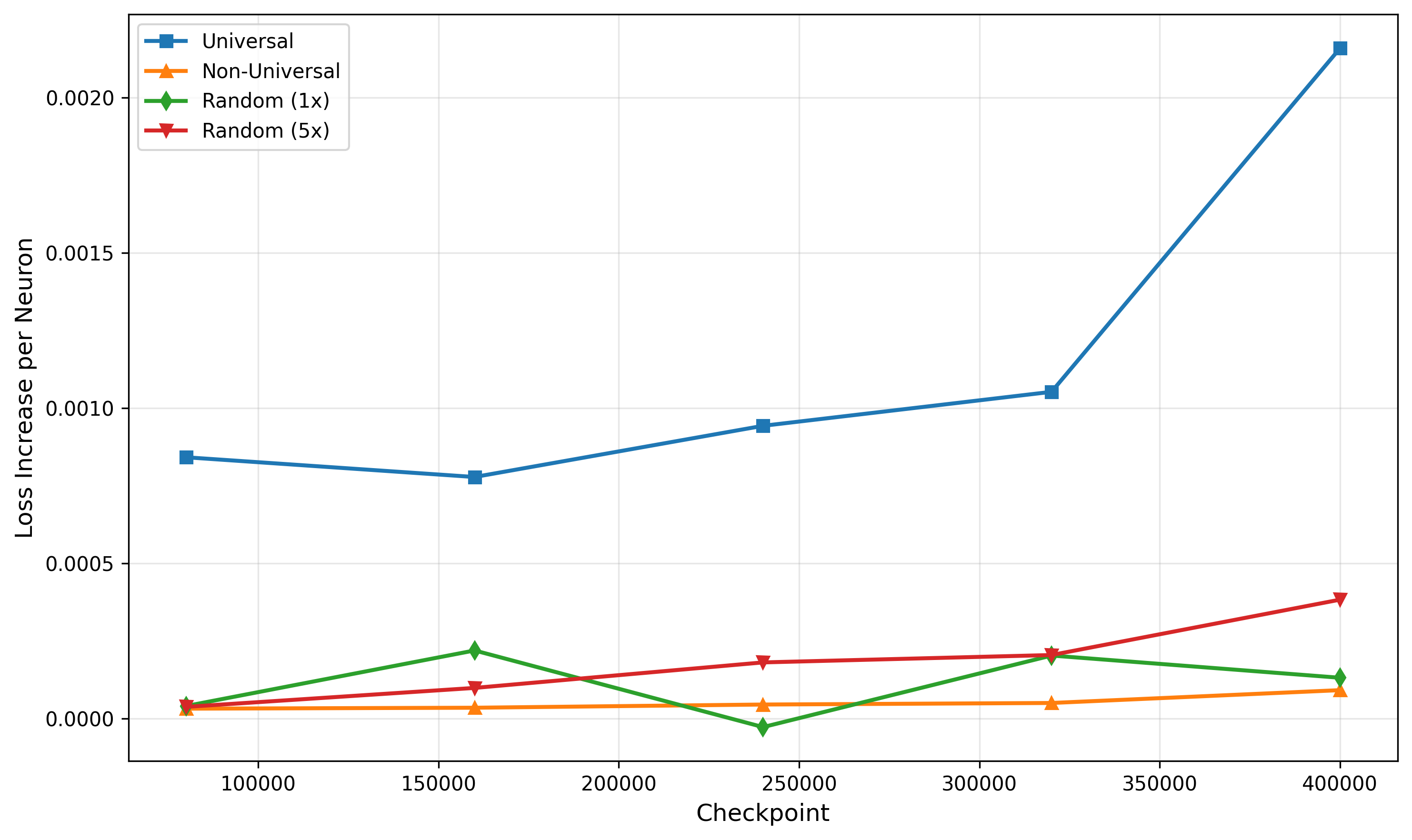}
    \caption{Ablation Efficiency (Change in loss per Neuron). The effect of universal neurons increases along with training steps. Compared to Nonuniversal neurons, they are crucial to the functionality of the language model.}
    \label{fig:ablate_eff}
  \end{minipage}
\end{figure}

We perform four types of ablation experiments: ablating all universal neurons, ablating all non-universal neurons, ablating a random set of neurons equal in number to the universal neurons, and ablating five times as many random neurons. Figures~\ref{fig:ablate_loss}\&\ref{fig:ablate_eff} show that ablating all universal neurons (with excess correlation $>0.5$) leads to a substantial negative impact in the model’s predictions compared to ablating random neurons. Ablating universal neurons causes around the same level of disruption as ablating 5 times as many random neurons.

\begin{figure}[t]
  \centering
  \begin{minipage}[t]{0.49\linewidth}
    \centering
    \includegraphics[width=\linewidth]{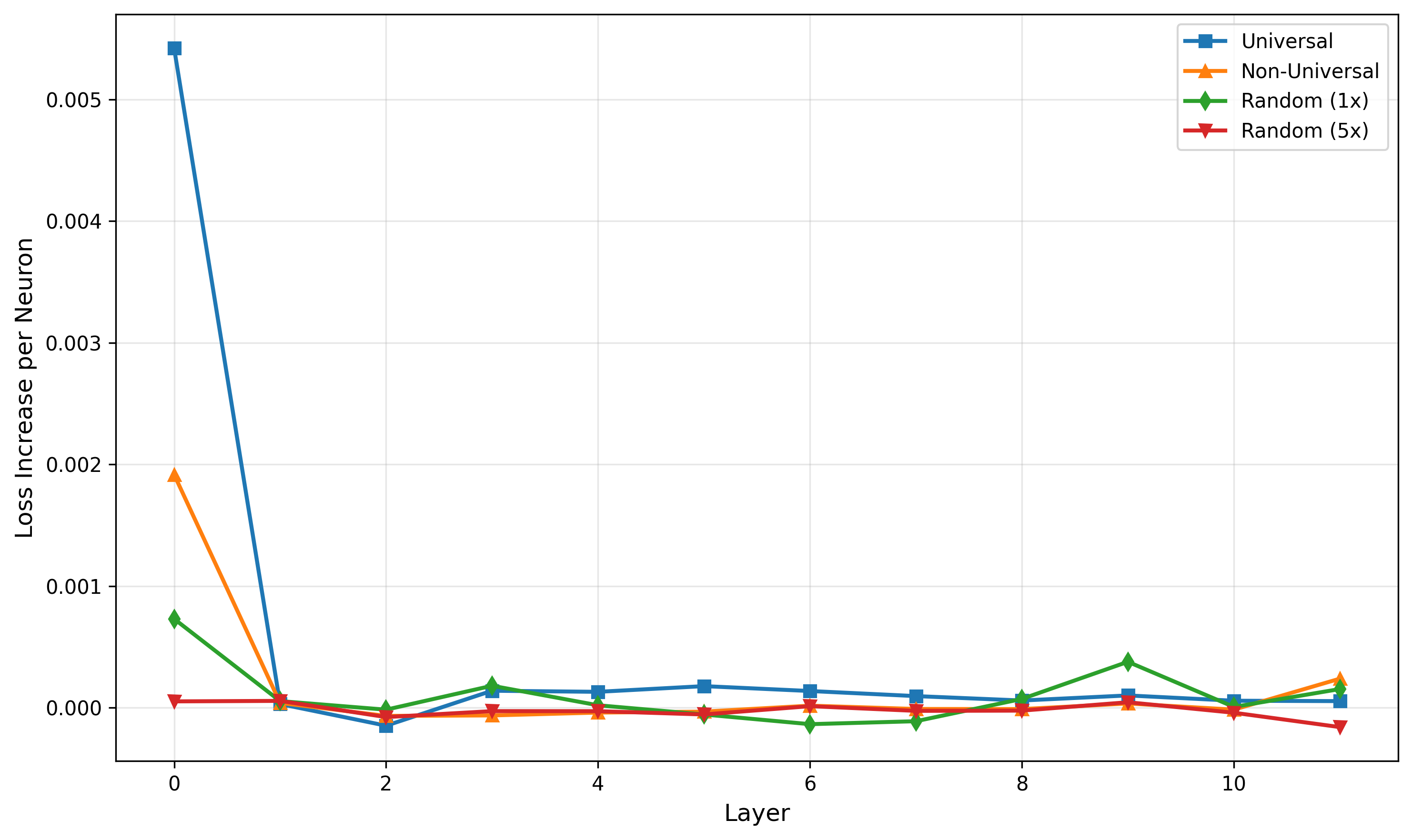}
    \caption{Layer-wise Ablation Efficiency (Change in loss per Neuron) on checkpoint 80k.}
    \label{fig:ablate_layer_eff_80}
  \end{minipage}%
  \hfill
  \begin{minipage}[t]{0.49\linewidth}
    \centering
    \includegraphics[width=\linewidth]{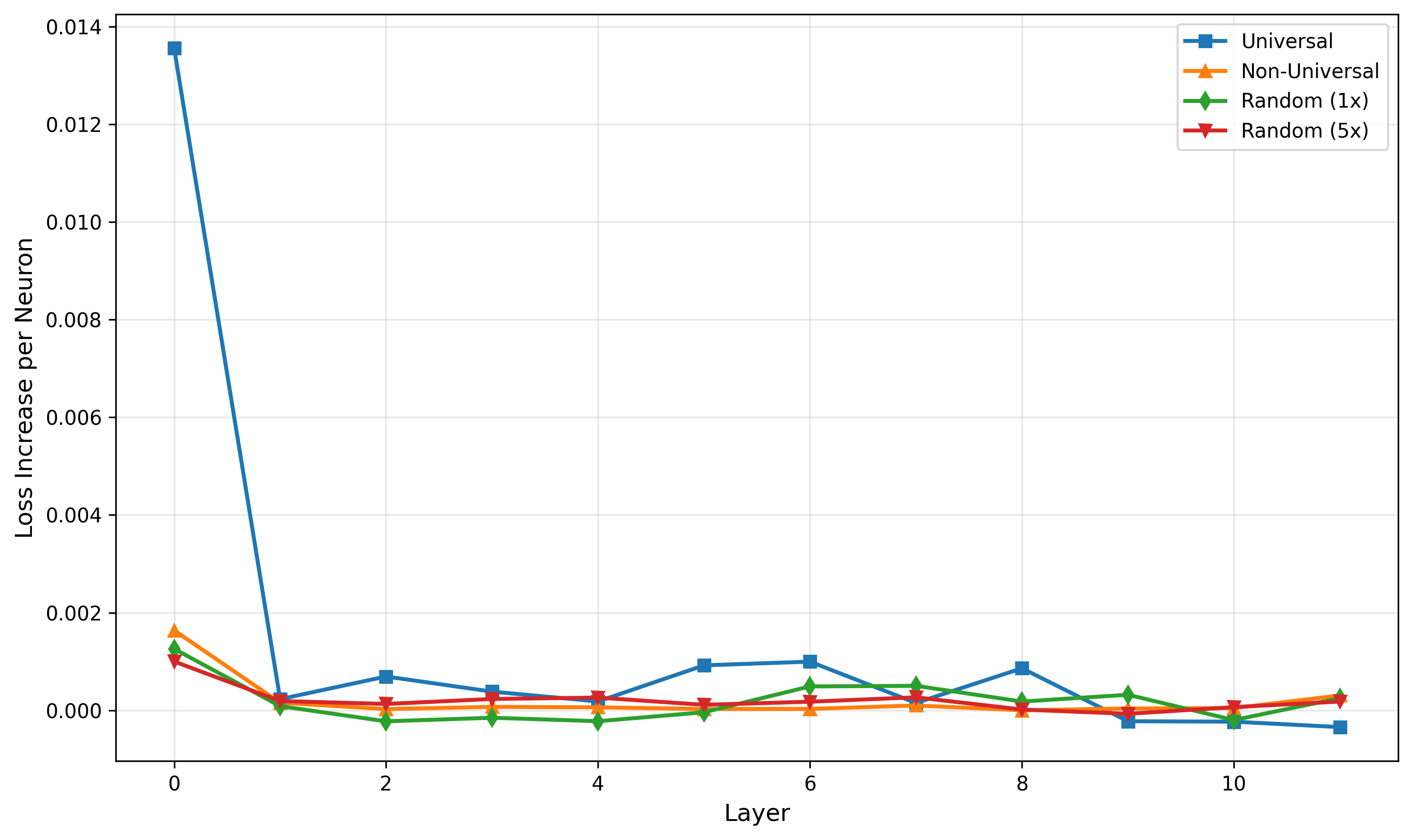}
    \caption{Layer-wise Ablation Efficiency (Change in loss per Neuron) on checkpoint 400k.}
    \label{fig:ablate_layer_eff_400}
  \end{minipage}
\end{figure}
These results demonstrate that universal neurons are not only shared across models but also causally important for inference. Considering that only about 4 to 5\% of all neurons are universal, our results strongly support their role as core components of the model’s learned algorithm.

In addition to the above findings, layer-wise ablation(Figures~\ref{fig:ablate_layer_eff_80}\&\ref{fig:ablate_layer_eff_400}) reveals that,throughout the training progress, ablating universal neurons in the first layer causes a disproportionately large increase in loss—far exceeding the impact observed in deeper layers. This suggests that early-layer universal neurons play a particularly critical role in shaping the model’s final predictions.

For completeness, we report comprehensive ablation experiments in Appendix~\ref{app:appC}\&\ref{app:appD}. These include global and layer-wise ablation results with different excess correlation thresholds (0.4 and 0.6). 

\section{Discussion and Conclusion}

\textbf{Findings} In this paper, we explore how universal neurons - neurons with high correlations across models - emerge early on in training and persist throughout checkpoints. We found that universal neurons have high functional significance, as ablating them results in higher loss per neuron than non-universal neuron ablations. These universal neurons remain consistent across training checkpoints, with both early and later layers having higher persistence on average. Trends in universality continue to remain stable despite threshold adjustment (0.4 and 0.6).

\textbf{Limitations} We only studied small models of a few hundred million parameters and monitored activations produced from a data subset of 5 million tokens, which is relatively small. Moreover, we only studied correlations between individual neurons as opposed to families of neurons or higher order circuits, which could offer more interpretable findings. 

\textbf{Future Work} To better understand the impact of universal neurons across families, it would be interesting to examine ablations for families of universal neurons and how the loss varies. A wider selection of experiments could lead to greater insight, for example, monitoring the effects of activation patching on some training data. 

\bibliographystyle{unsrt}
\bibliography{references}

\appendix
\section{Layer-wise Universal Neuron Percentage for Different Thresholds}
\label{app:appA}
Changing the threshold for labeling universal neuron shows similar trends.
\begin{figure}[htbp]
  \centering
  \begin{minipage}[t]{0.49\linewidth}
    \centering
    \includegraphics[width=\linewidth]{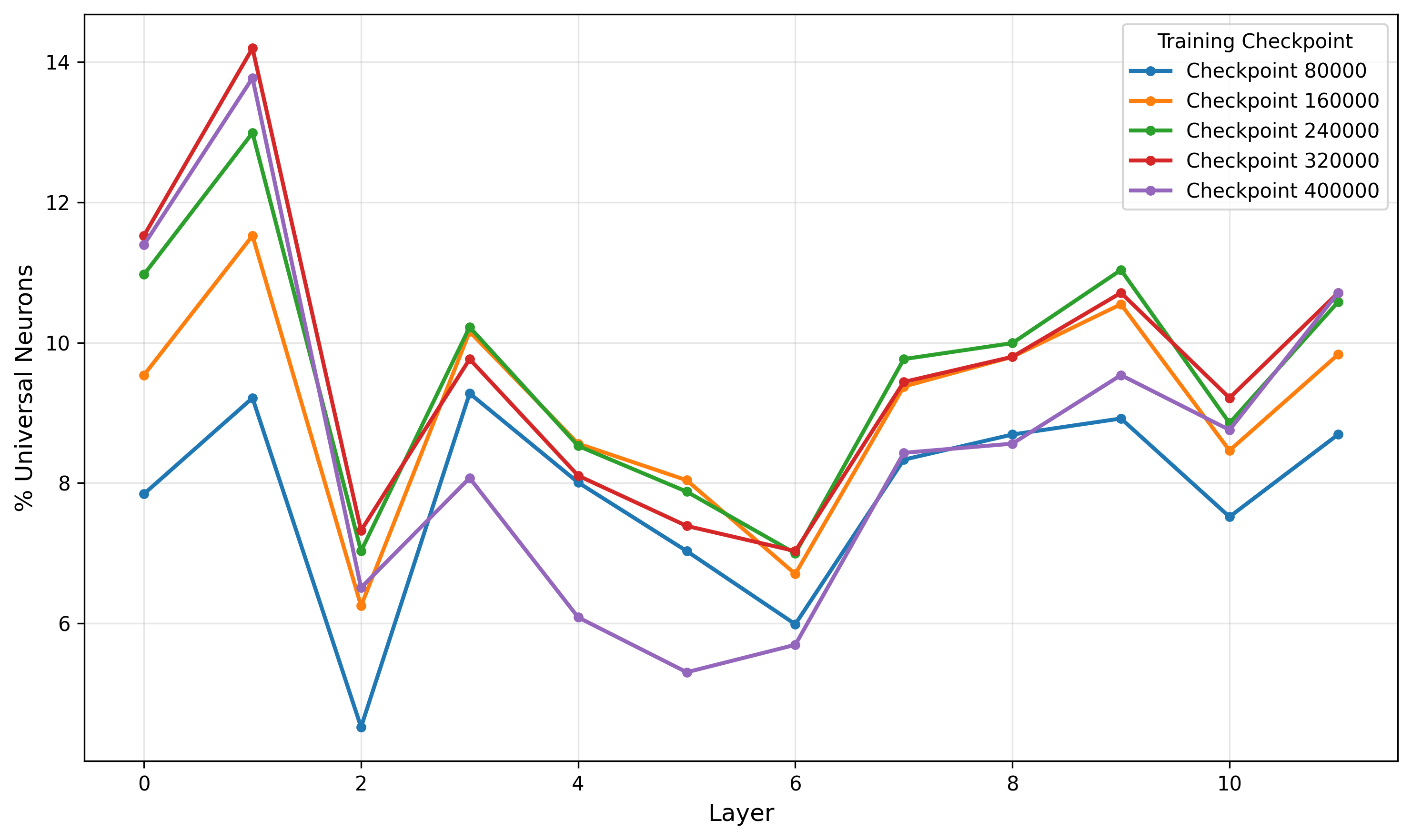}
    \caption{Percentage of Universal Neurons(Excess correlation > 0.4) Across Layers throughout different checkpoints.}
  \end{minipage}%
  \hfill
  \begin{minipage}[t]{0.49\linewidth}
    \centering
    \includegraphics[width=\linewidth]{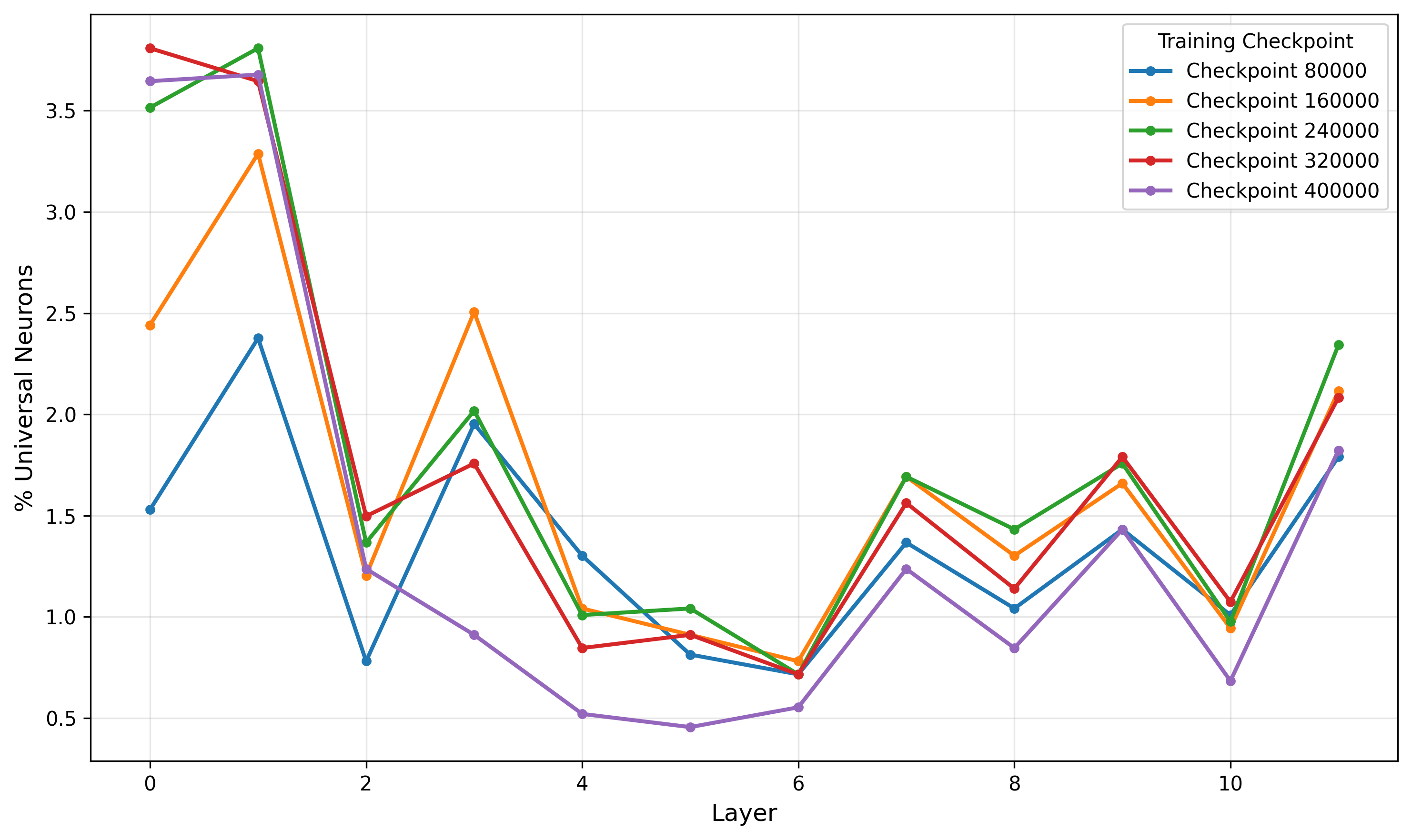}
    \caption{Percentage of Universal Neurons(Excess correlation > 0.6) Across Layers throughout different checkpoints.}
  \end{minipage}
\end{figure}

\section{Persistence of Universal Neurons over Checkpoints for Different Thresholds}
All thresholds exhibit a U-shaped trend: lower persistence in middle layers (3–8) and higher stability in both early and especially late layers. This suggests that early and late layers encode more stable, model-aligned features during training.
\label{app:appB}
\begin{figure}[htbp]
  \centering
  \begin{minipage}[t]{0.49\linewidth}
    \centering
    \includegraphics[width=\linewidth]{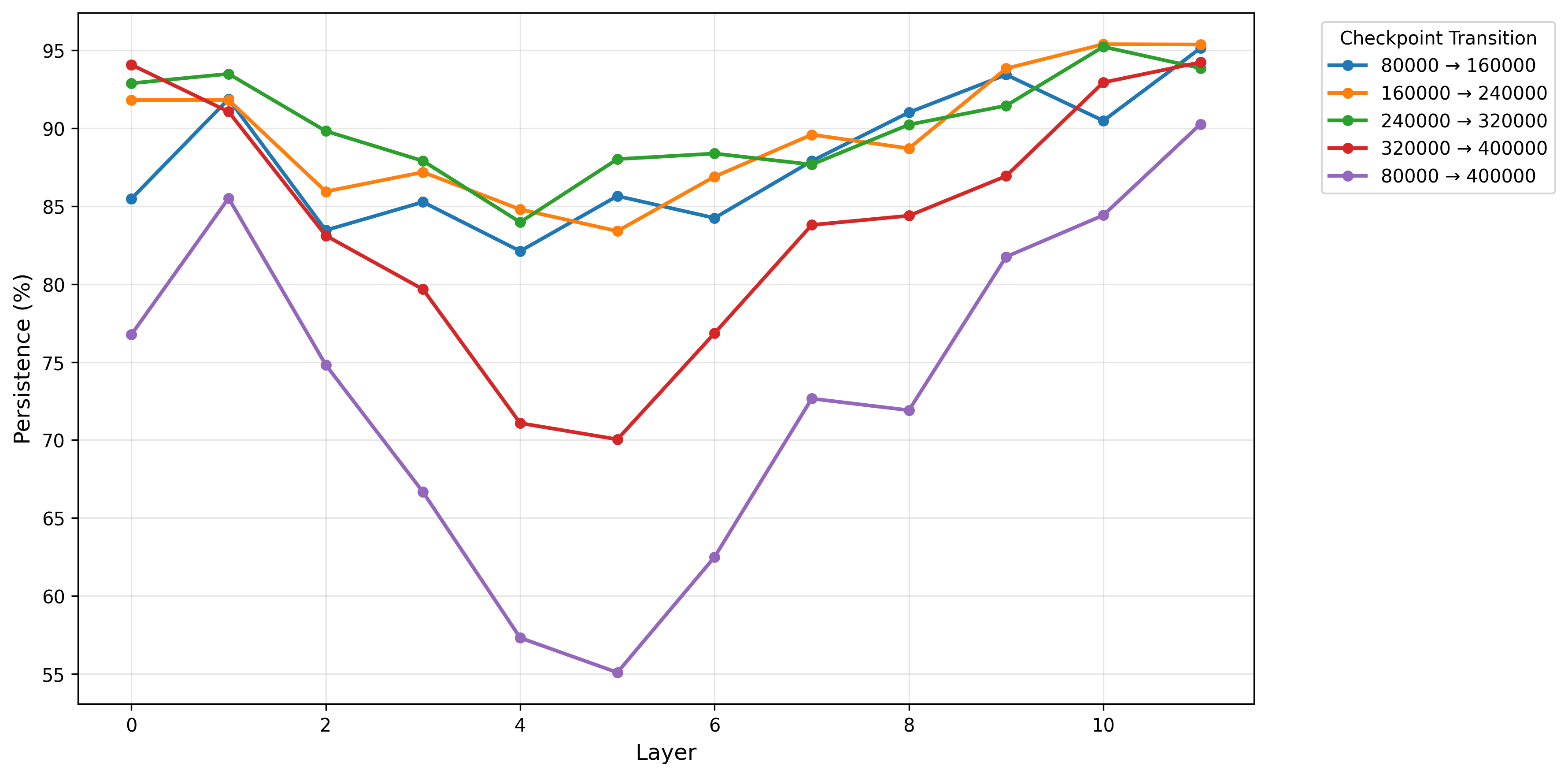}
    \caption{Persistence of Universal Neurons (Excess correlation > 0.4) Across Layers throughout different checkpoints.}
  \end{minipage}%
  \hfill
  \begin{minipage}[t]{0.49\linewidth}
    \centering
    \includegraphics[width=\linewidth]{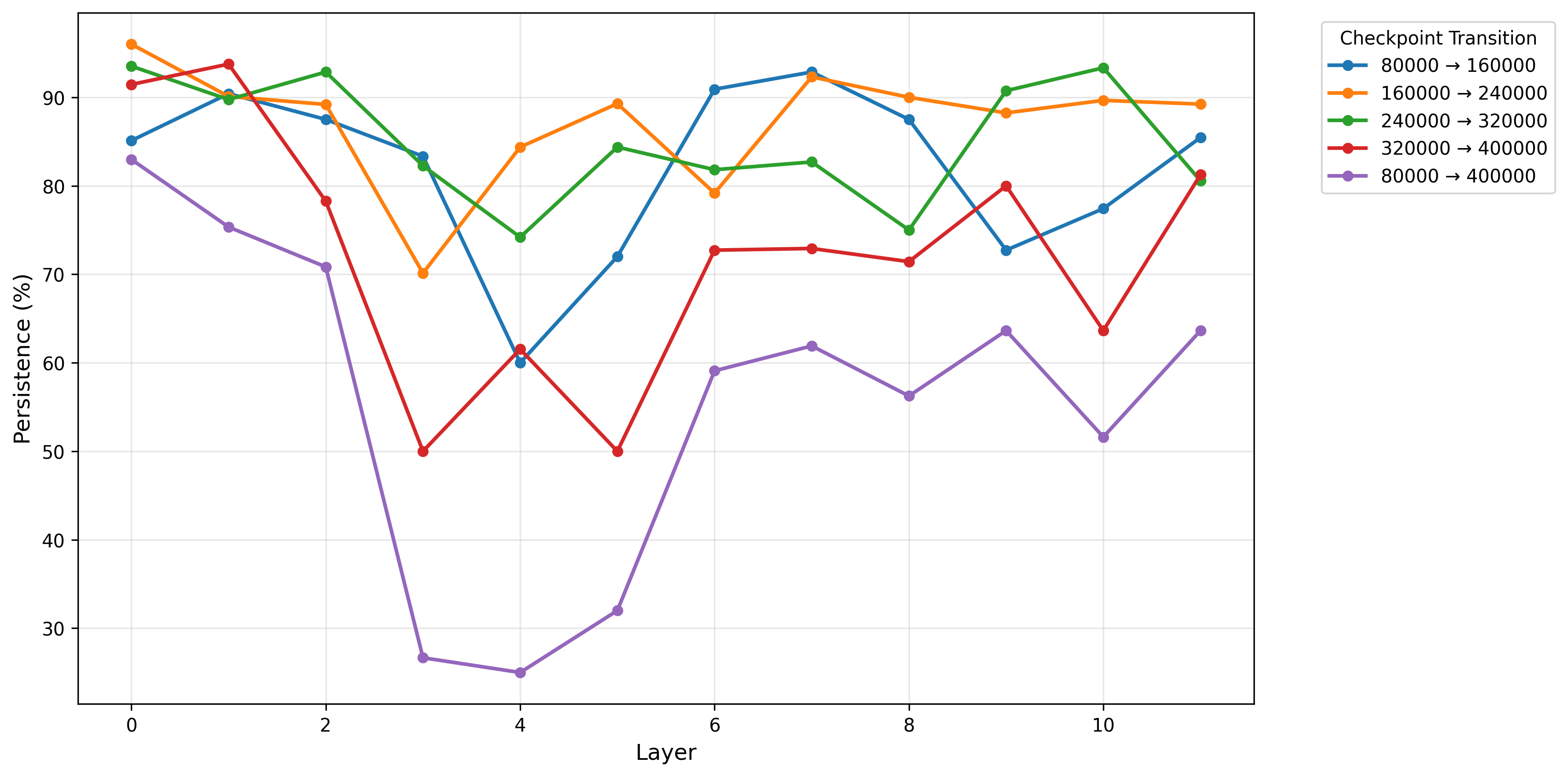}
    \caption{Persistence of Universal Neurons (Excess correlation > 0.6) Across Layers throughout different checkpoints.}
  \end{minipage}
\end{figure}
\clearpage

\section{Additional Universal Neuron Ablation Experiment Results}
\label{app:appC}
\begin{figure}[htbp]
  \centering
  \begin{minipage}[t]{0.49\linewidth}
    \centering
    \includegraphics[width=\linewidth]{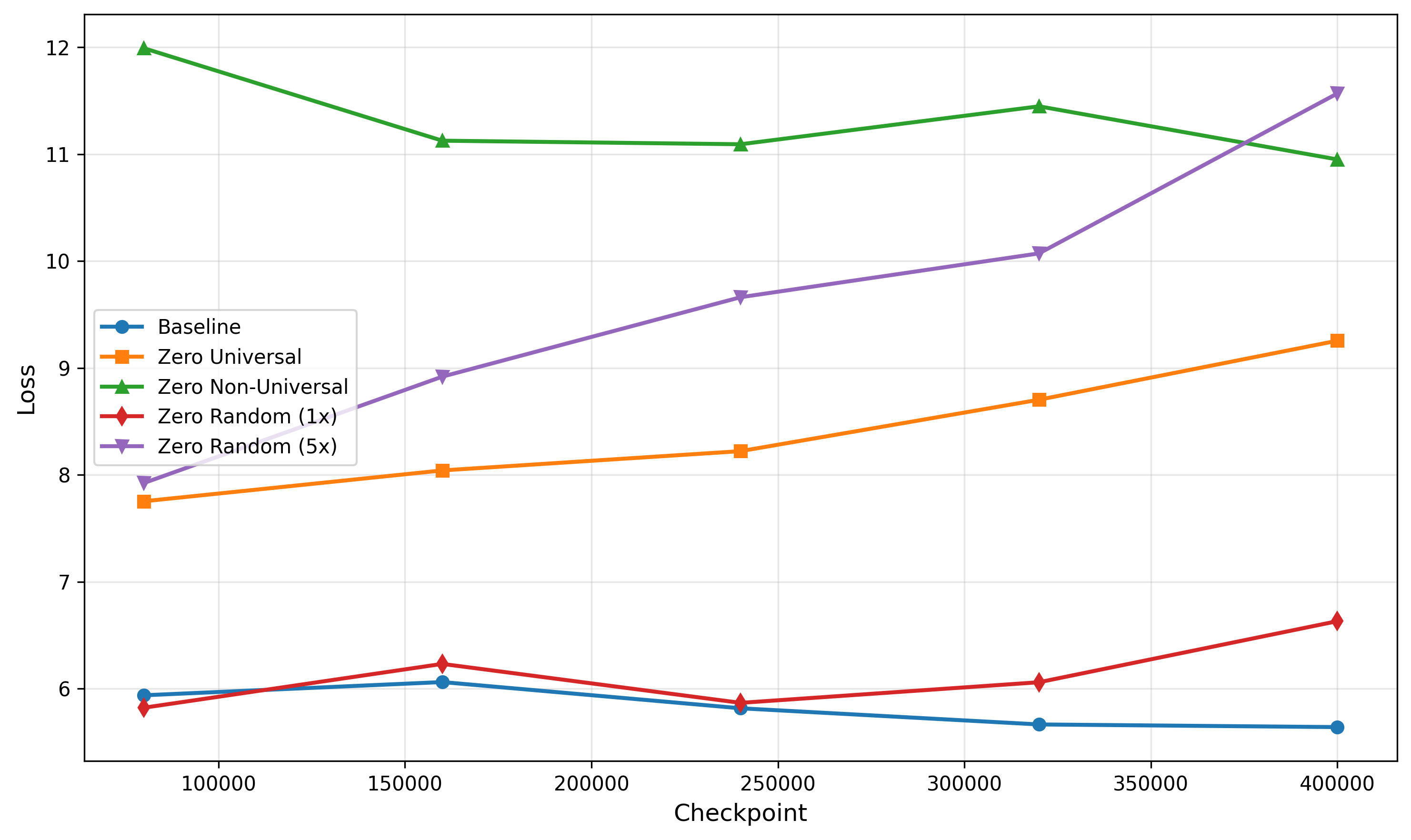}
    \caption{Absolute Loss Values After Ablating(zeroing activations) Different Neurons. When we lower the threshold from an excess correlation of 0.5 to that of 0.4, ablating 5x the amount of random neurons achieve more disruptive results than ablating all universal neurons.}
    \label{fig:ablate_loss_04}
  \end{minipage}%
  \hfill
  \begin{minipage}[t]{0.49\linewidth}
    \centering
    \includegraphics[width=\linewidth]{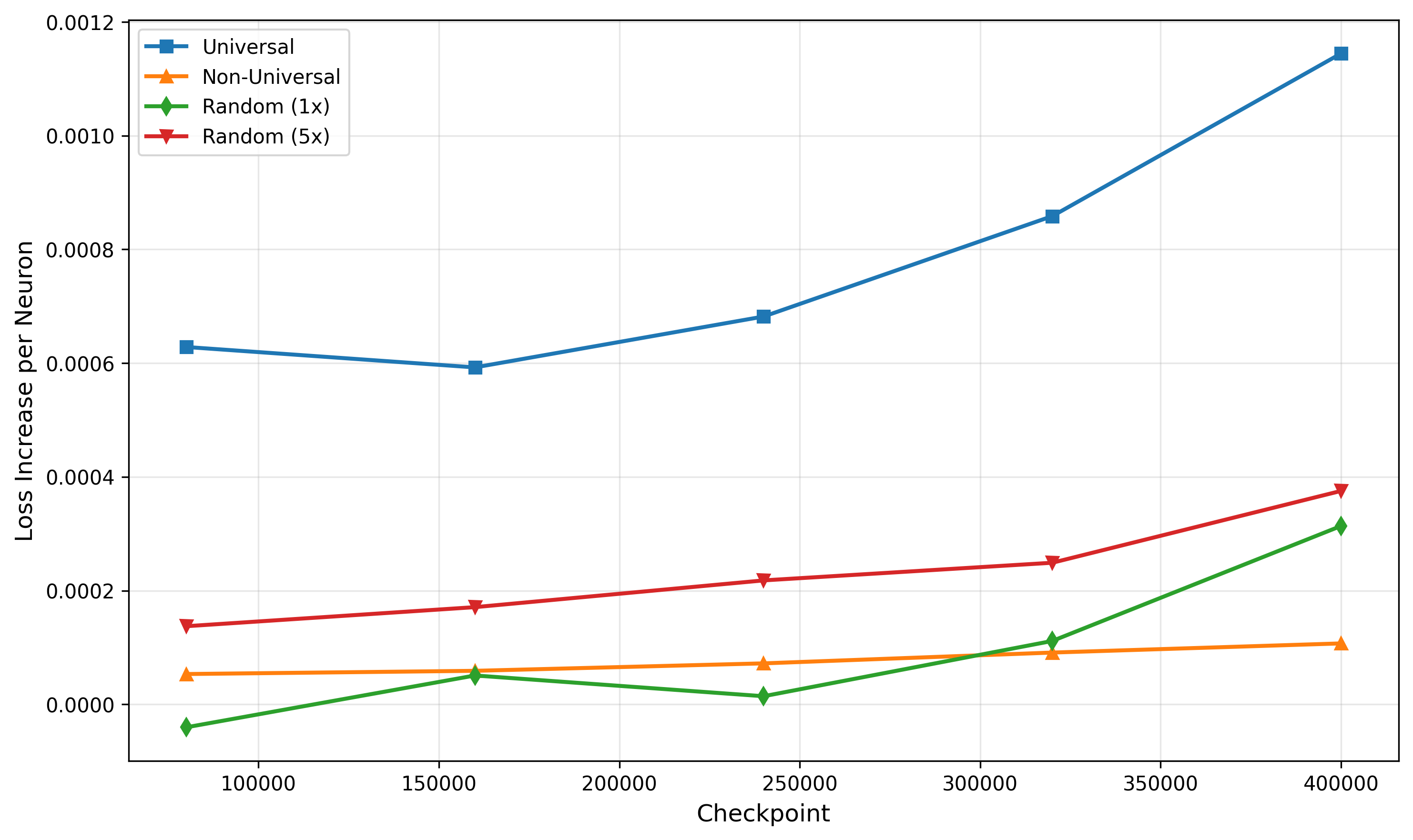}
    \caption{Ablation Efficiency (Change in loss per Neuron). The effect of universal neurons increases along with training steps. Compared to Nonuniversal neurons, they are crucial to the functionality of the language model.}
    \label{fig:ablate_eff_04}
  \end{minipage}
\end{figure}

\begin{figure}[htbp]
  \centering
  \begin{minipage}[t]{0.49\linewidth}
    \centering
    \includegraphics[width=\linewidth]{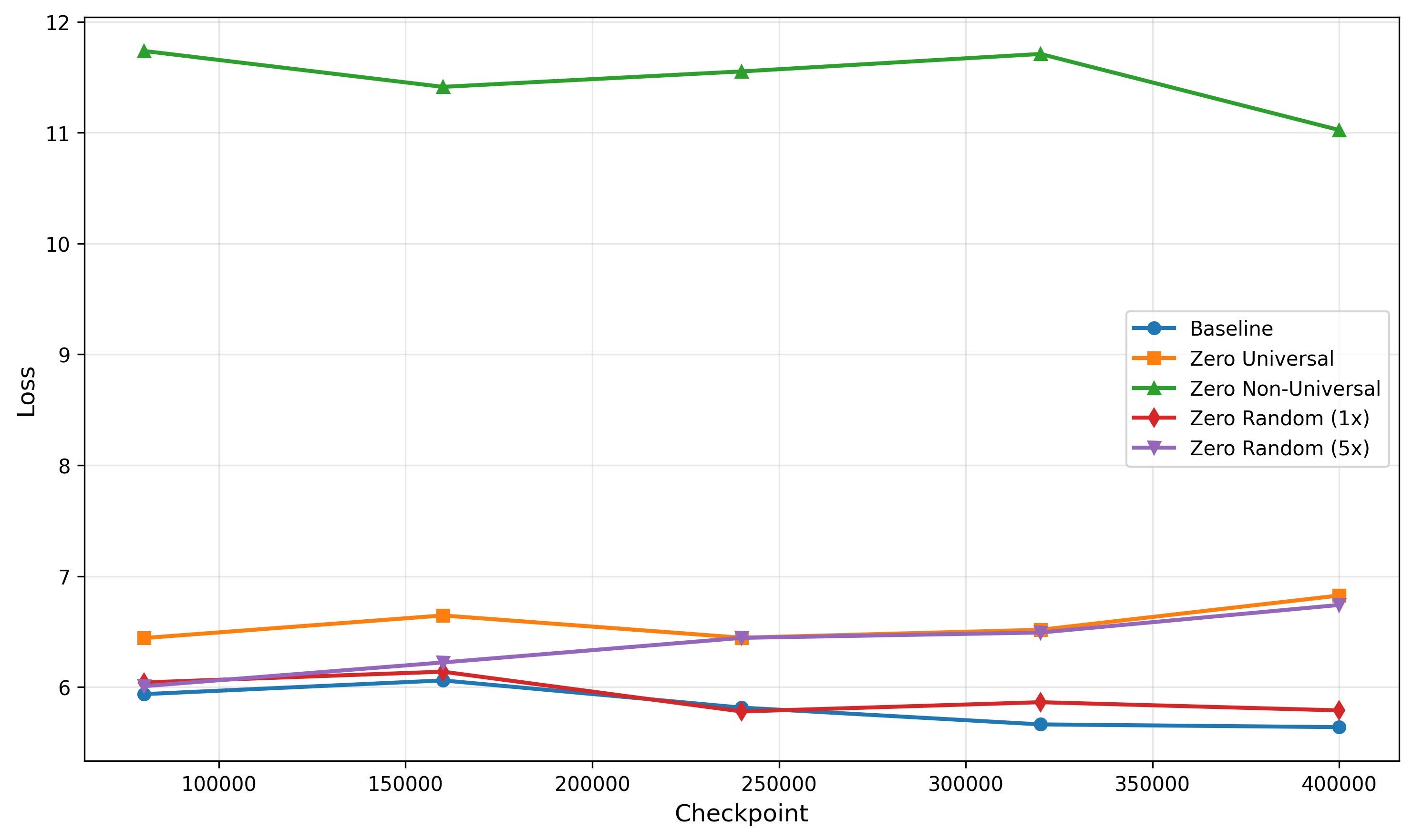}
    \caption{Absolute Loss Values After Ablating(zeroing activations) Different Neurons. It takes around 5x the amount of random neurons to achieve the same disruptive result of ablating universal neurons.}
    \label{fig:ablate_loss_06}
  \end{minipage}%
  \hfill
  \begin{minipage}[t]{0.49\linewidth}
    \centering
    \includegraphics[width=\linewidth]{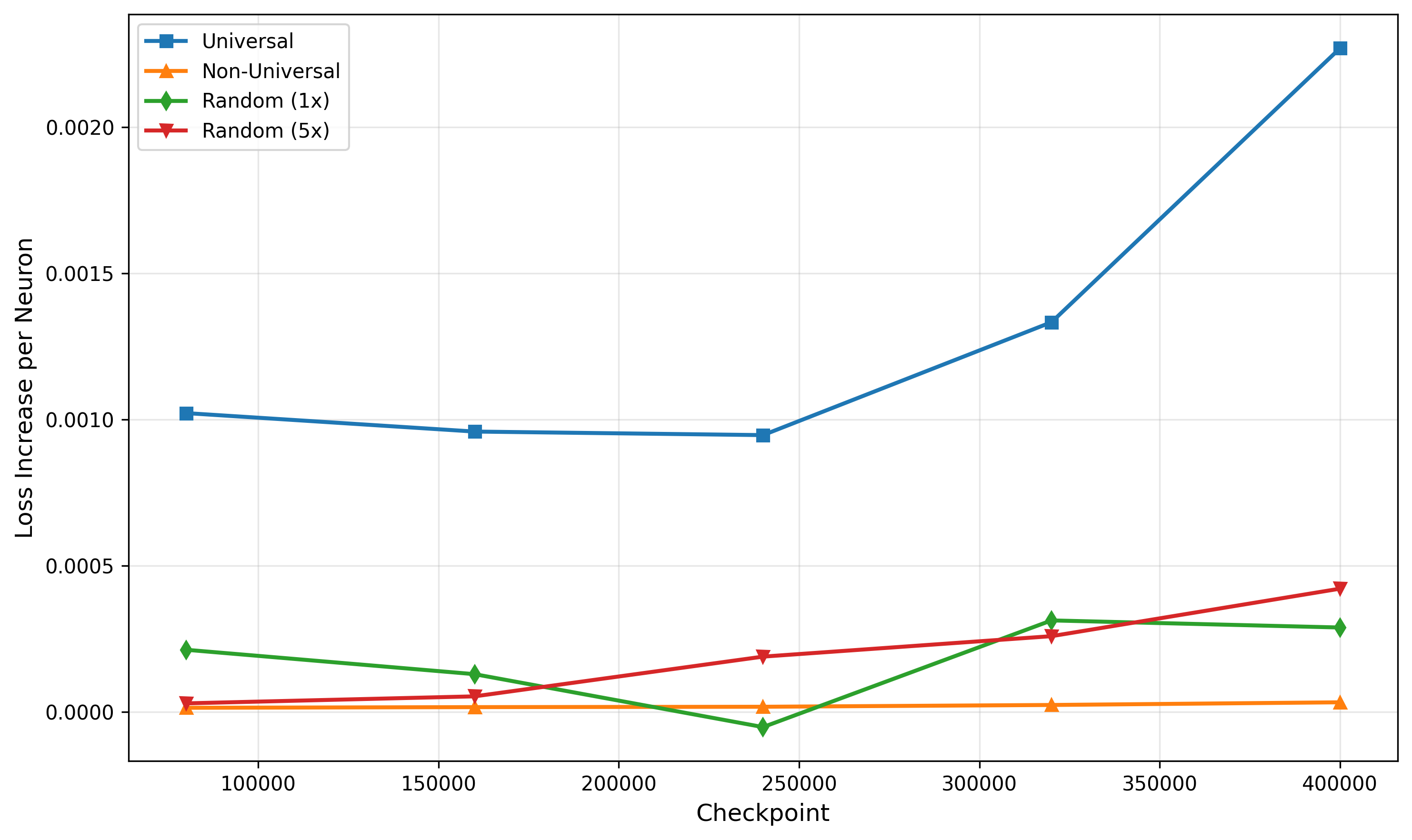}
    \caption{Ablation Efficiency (Change in loss per Neuron). The effect of universal neurons increases along with training steps. Compared to Nonuniversal neurons, they are crucial to the functionality of the language model.}
    \label{fig:ablate_eff_06}
  \end{minipage}
\end{figure}

As confirmed in Figure \ref{fig:ablate_loss_04}\&\ref{fig:ablate_loss_06}, universal neurons are more significant than nonuniversal neurons, but the magnitude differed greatly. The impact of ablating universal neurons when thresholding with 0.4 is greatly overwhelmed by ablating 5x the amount of random neurons. Whereas the impact of ablating universal neurons when thresholding with 0.6 is comparable to that of ablating the same number of random neurons. 

However, Figure \ref{fig:ablate_eff_04}\&\ref{fig:ablate_eff_06} consistently prove the functional significance of universal neurons through per-neuron metric.

\subsection{Ablation Loss Increase with 0.4 Excess Correlation Thresholding}

\subsection{Ablation Loss Increase with 0.6 Excess Correlation Thresholding}

\clearpage
\section{Additional Layer-Wise Universal Neuron Ablation Experiment Results}
\label{app:appD}
As shown in all figures below, the first layer of the model contains universal neurons that are the most crucial. 

\subsection{Layer-wise Ablation Efficiency Across Checkpoints}
\begin{figure}[htbp]
  \centering
  \begin{minipage}[t]{0.49\linewidth}
    \centering
    \includegraphics[width=\linewidth]{new_vis/layerwise_efficiency_checkpoint_80000.png}
    \caption{Layer-wise Ablation Efficiency (Change in loss per Neuron) on checkpoint 80k.}
  \end{minipage}%
  \hfill
  \begin{minipage}[t]{0.49\linewidth}
    \centering
    \includegraphics[width=\linewidth]{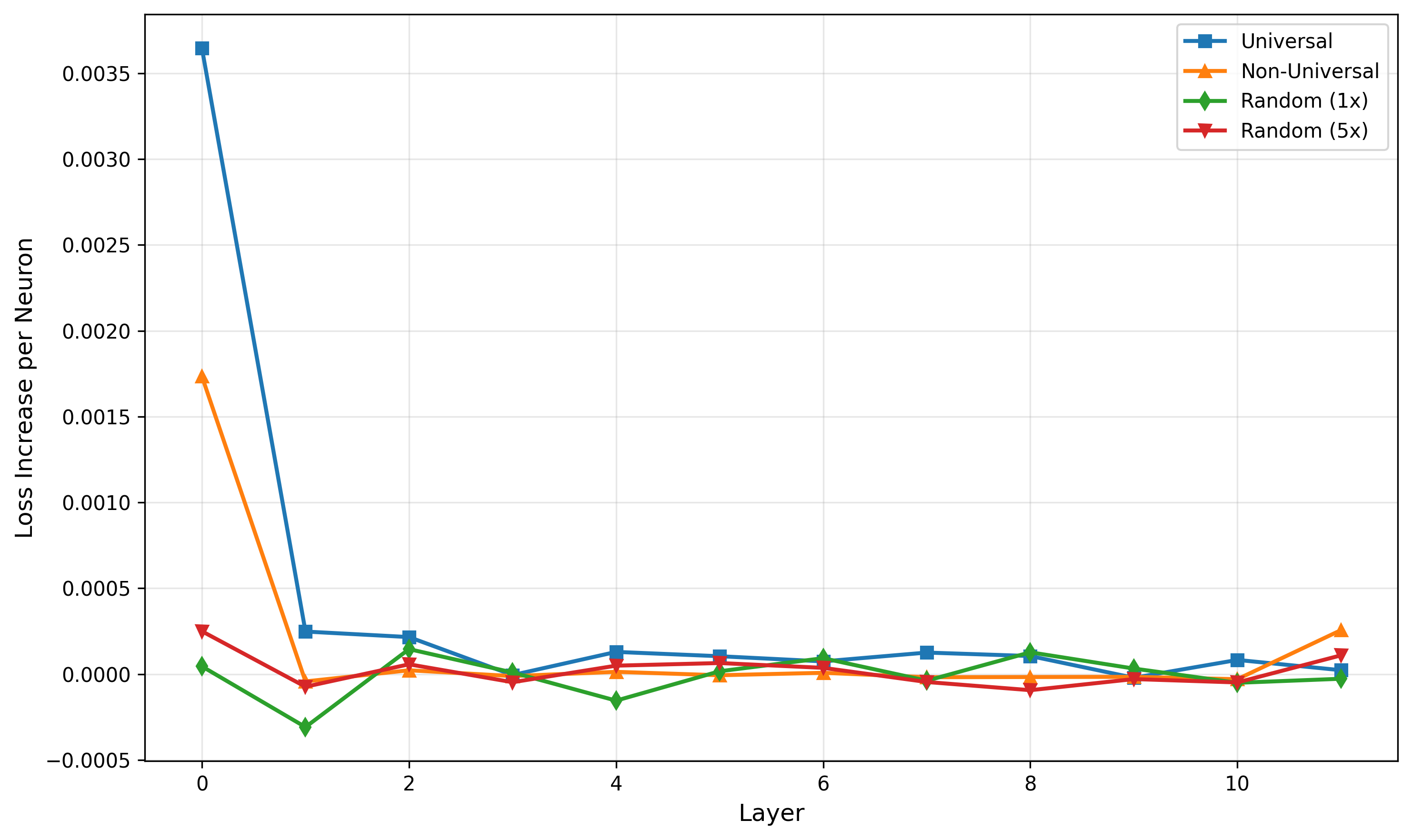}
    \caption{Layer-wise Ablation Efficiency (Change in loss per Neuron) on checkpoint 160k.}
  \end{minipage}
\end{figure}
\begin{figure}[htbp]
  \centering
  \begin{minipage}[t]{0.49\linewidth}
    \centering
    \includegraphics[width=\linewidth]{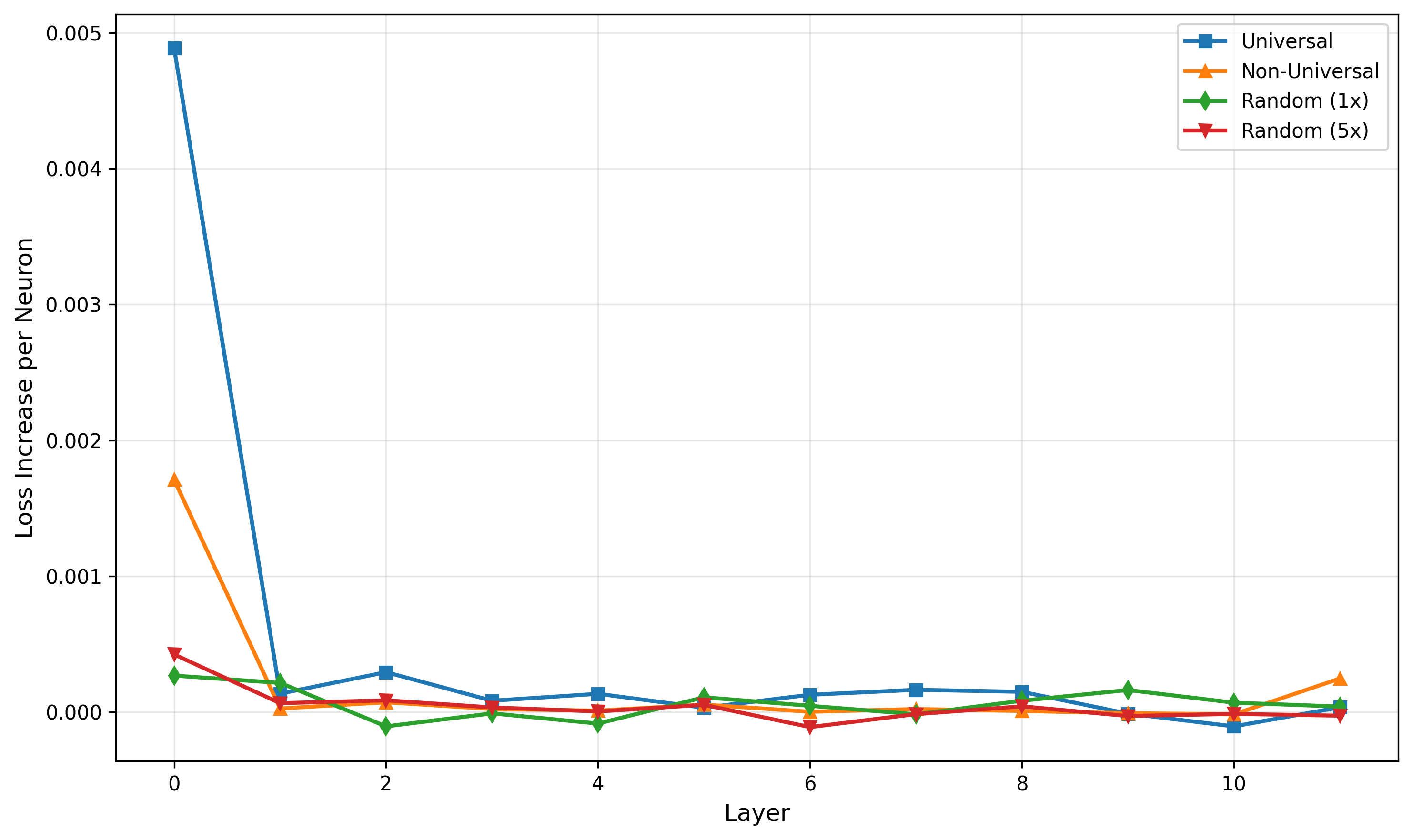}
    \caption{Layer-wise Ablation Efficiency (Change in loss per Neuron) on checkpoint 240k.}
  \end{minipage}%
  \hfill
  \begin{minipage}[t]{0.49\linewidth}
    \centering
    \includegraphics[width=\linewidth]{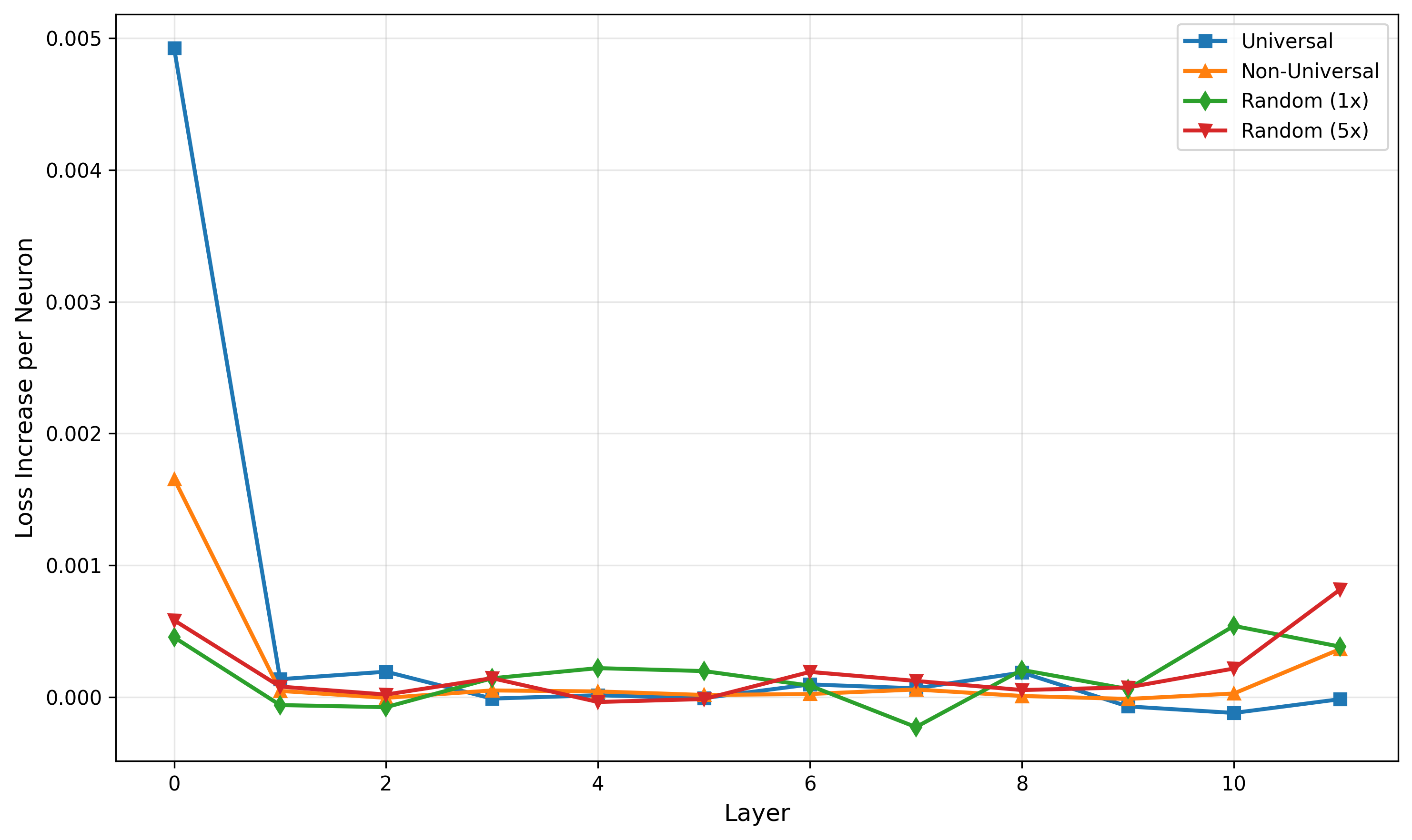}
    \caption{Layer-wise Ablation Efficiency (Change in loss per Neuron) on checkpoint 320k.}
  \end{minipage}
\end{figure}
\begin{figure}[htbp]
  \centering
    \includegraphics[width=0.49\linewidth]{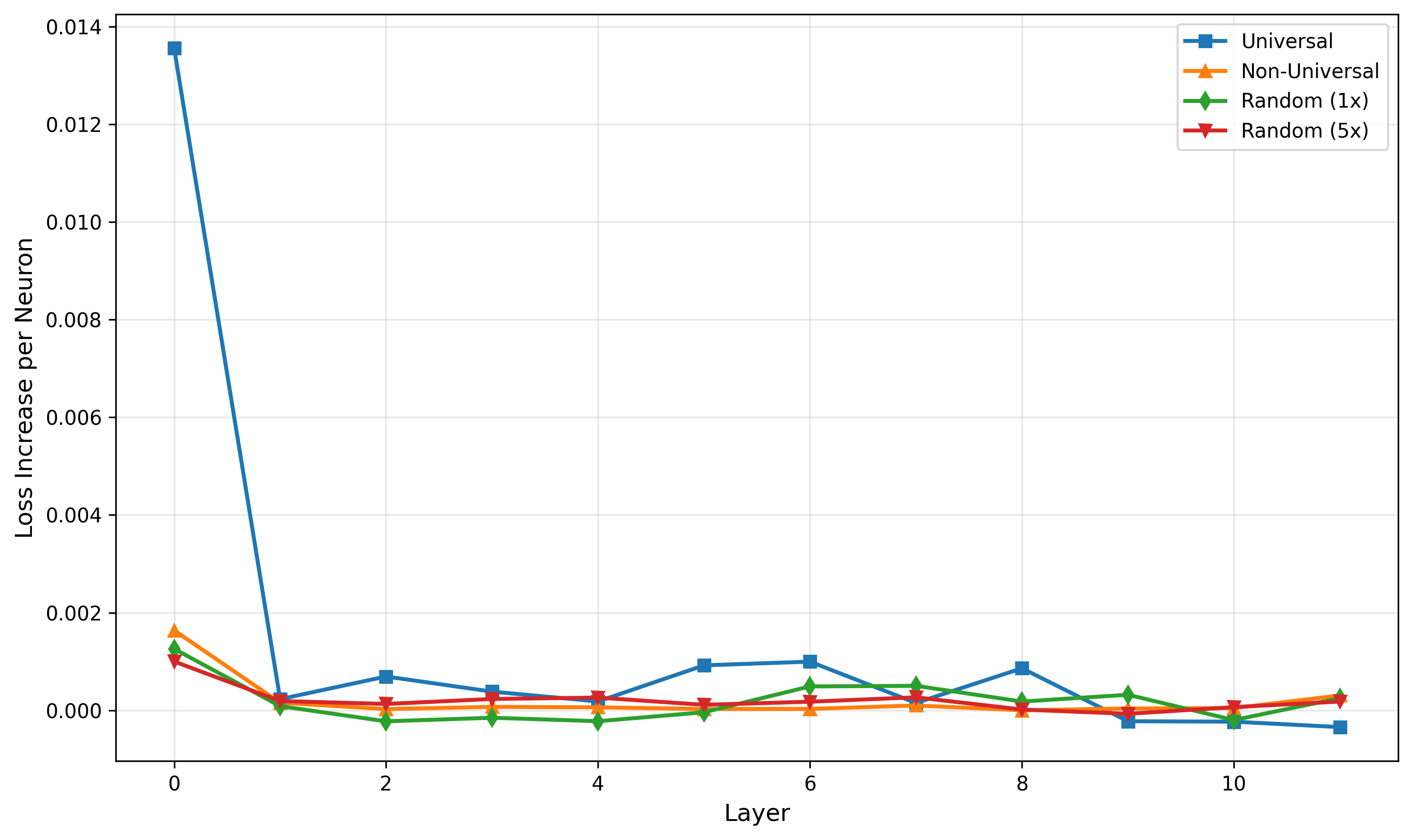}
    \caption{Layer-wise Ablation Efficiency (Change in loss per Neuron) on checkpoint 400k.}
\end{figure}
\clearpage
\subsection{Layer-wise Ablation Efficiency with 0.4 Excess Correlation Thresholding}
\begin{figure}[htbp]
  \centering
  \begin{minipage}[t]{0.49\linewidth}
    \centering
    \includegraphics[width=\linewidth]{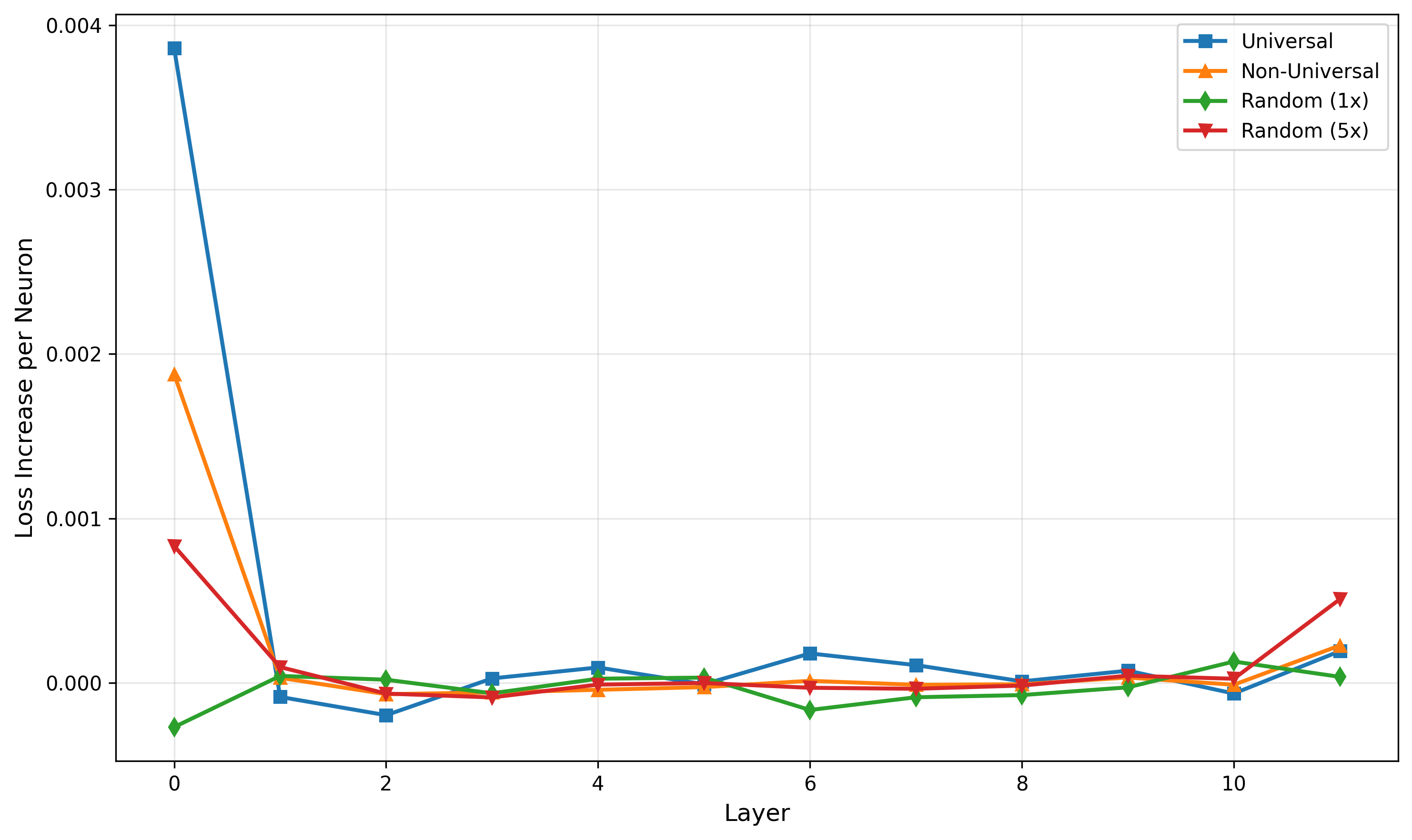}
    \caption{Layer-wise Ablation Efficiency (Change in loss per Neuron) on checkpoint 80k.}
  \end{minipage}%
  \hfill
  \begin{minipage}[t]{0.49\linewidth}
    \centering
    \includegraphics[width=\linewidth]{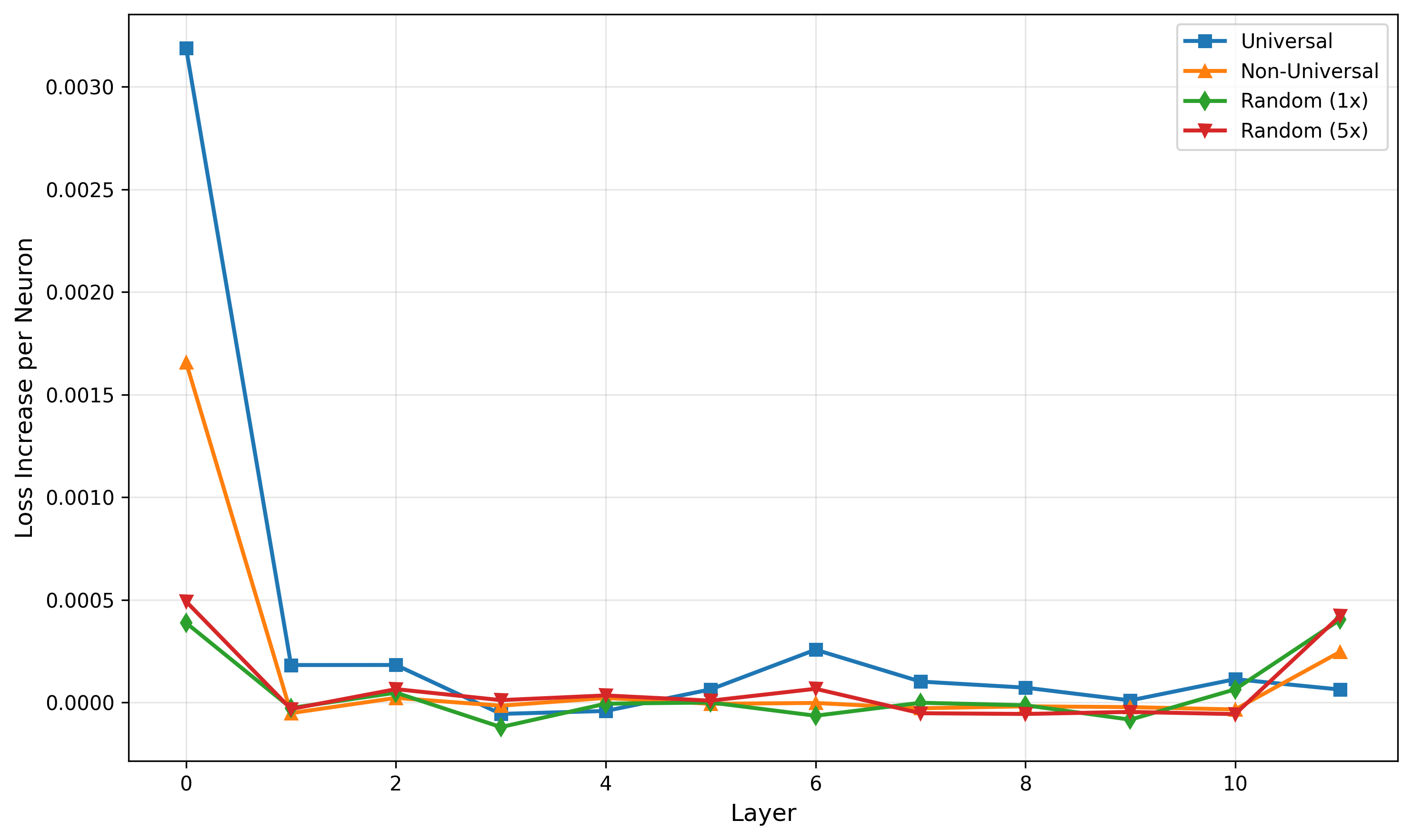}
    \caption{Layer-wise Ablation Efficiency (Change in loss per Neuron) on checkpoint 160k.}
  \end{minipage}
\end{figure}
\begin{figure}[htbp]
  \centering
  \begin{minipage}[t]{0.49\linewidth}
    \centering
    \includegraphics[width=\linewidth]{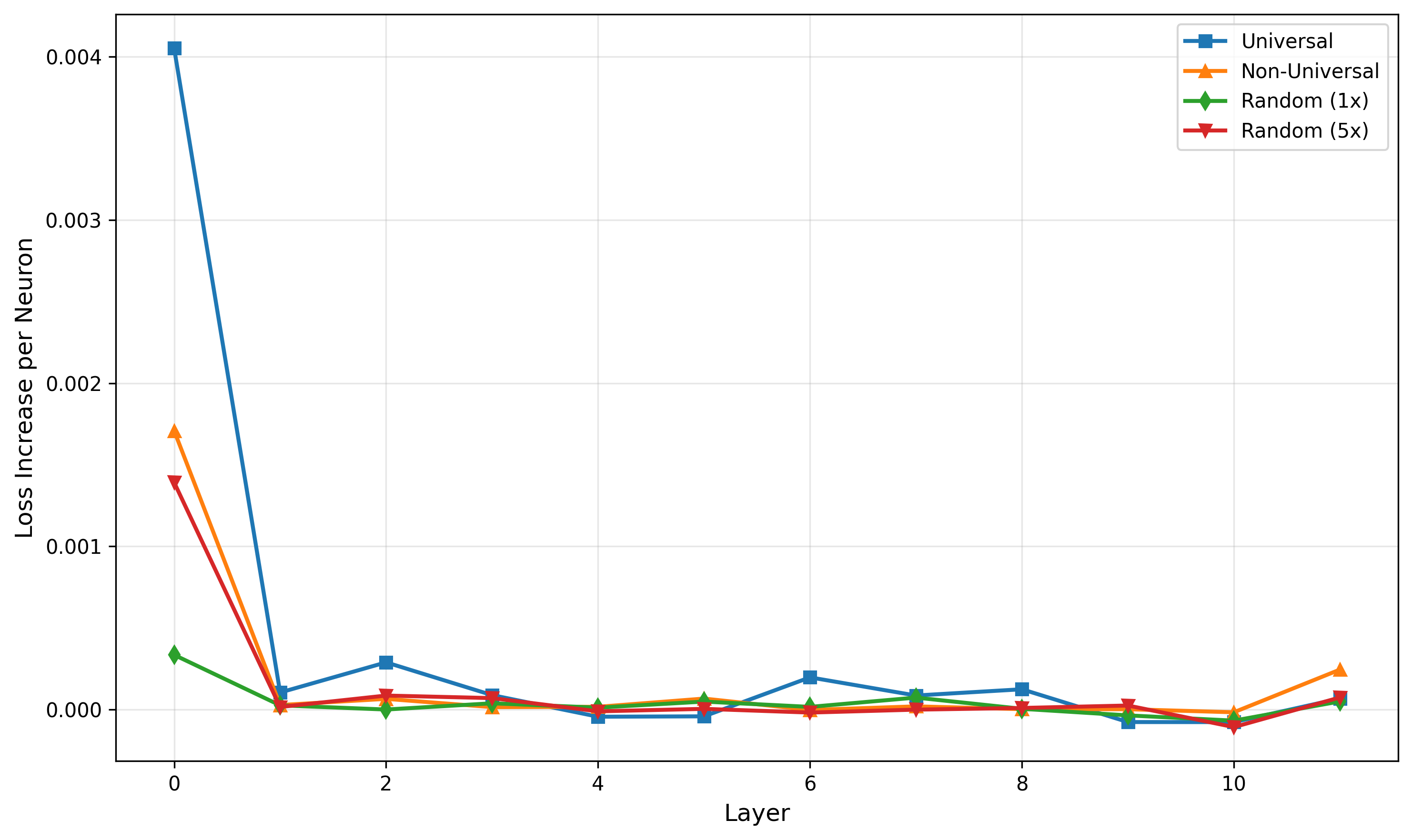}
    \caption{Layer-wise Ablation Efficiency (Change in loss per Neuron) on checkpoint 240k.}
  \end{minipage}%
  \hfill
  \begin{minipage}[t]{0.49\linewidth}
    \centering
    \includegraphics[width=\linewidth]{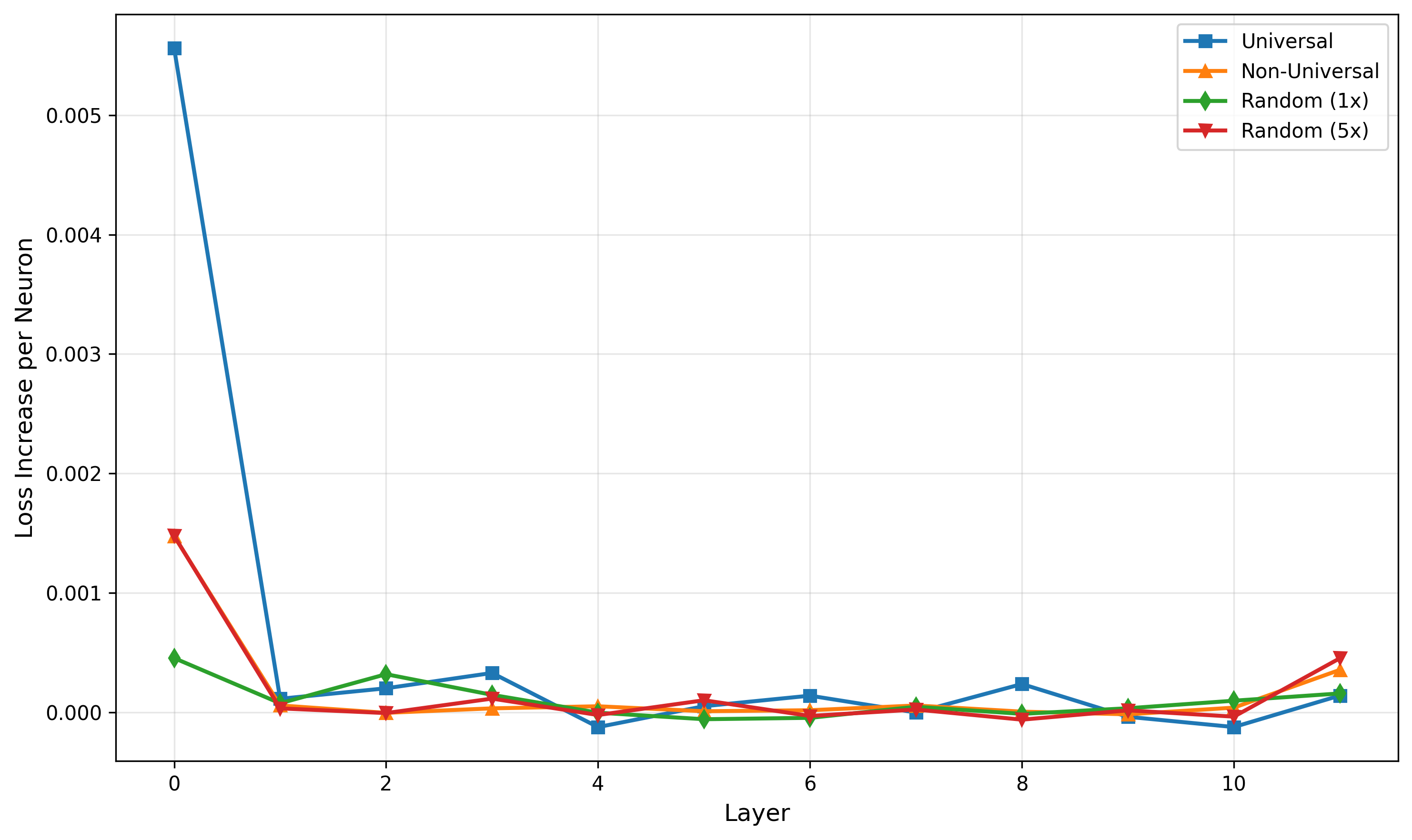}
    \caption{Layer-wise Ablation Efficiency (Change in loss per Neuron) on checkpoint 320k.}
  \end{minipage}
\end{figure}
\begin{figure}[htbp]
  \centering
    \includegraphics[width=0.49\linewidth]{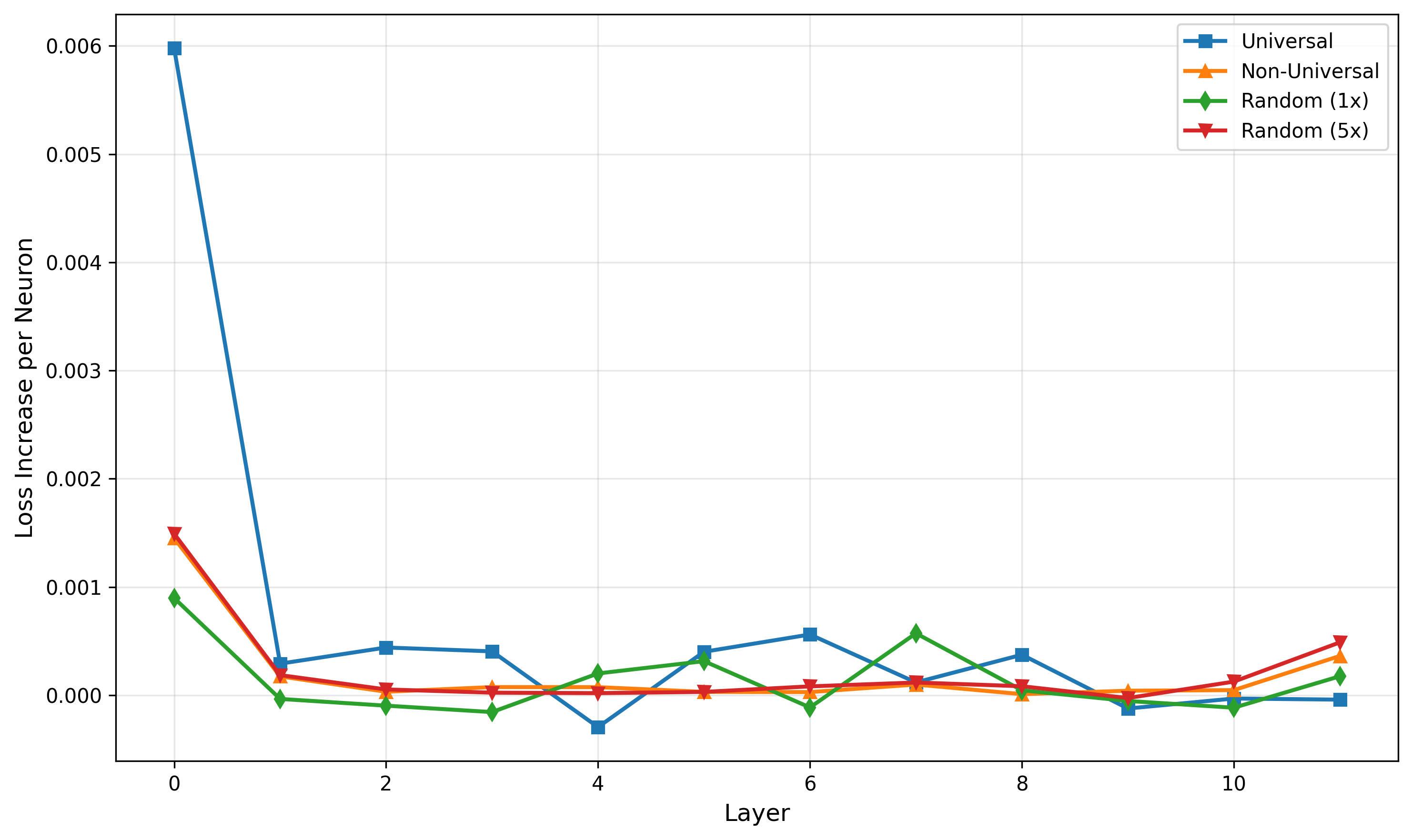}
    \caption{Layer-wise Ablation Efficiency (Change in loss per Neuron) on checkpoint 400k.}
\end{figure}

\clearpage
\subsection{Layer-wise Ablation Efficiency with 0.6 Excess Correlation Thresholding}
\begin{figure}[htbp]
  \centering
  \begin{minipage}[t]{0.49\linewidth}
    \centering
    \includegraphics[width=\linewidth]{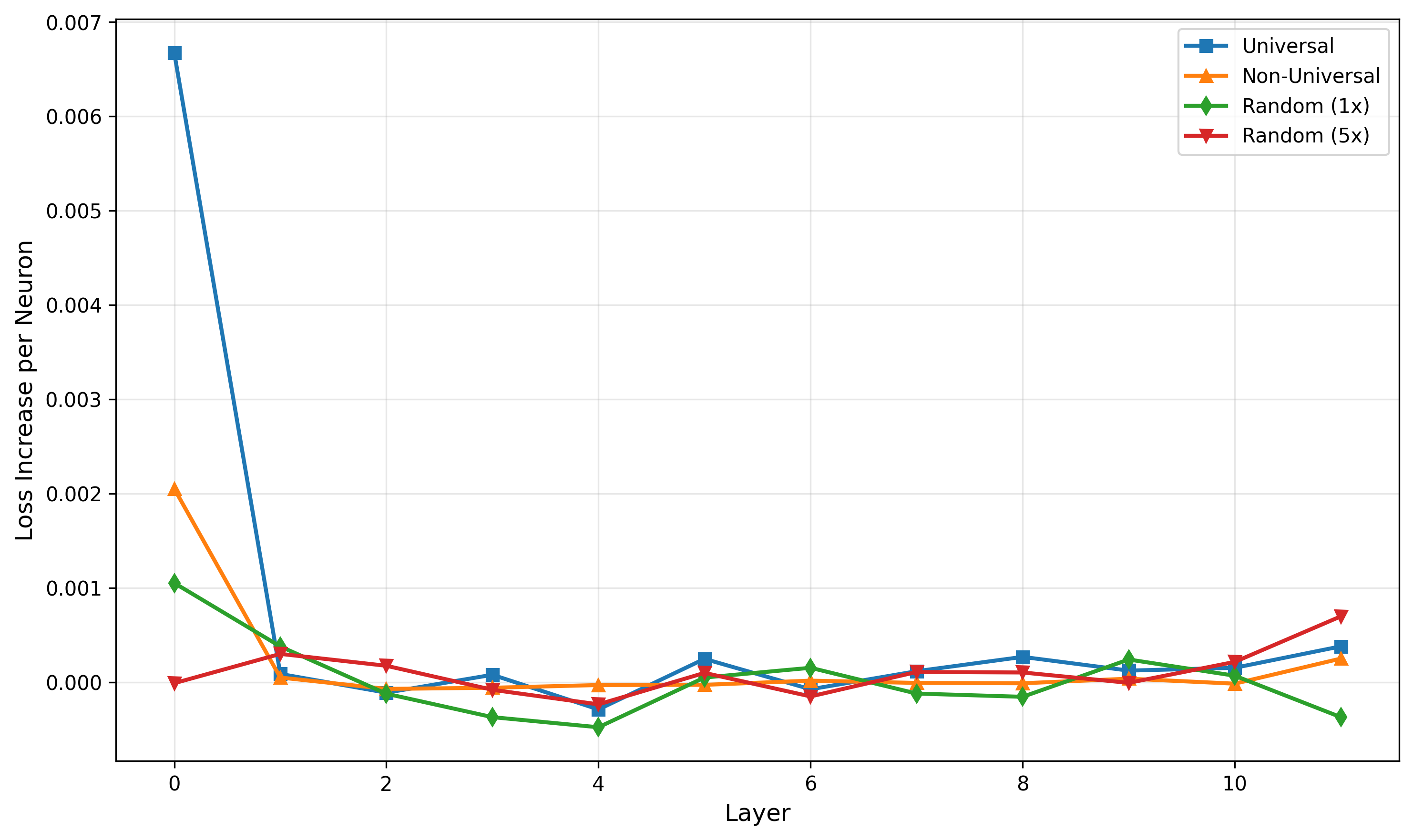}
    \caption{Layer-wise Ablation Efficiency (Change in loss per Neuron) on checkpoint 80k.}
  \end{minipage}%
  \hfill
  \begin{minipage}[t]{0.49\linewidth}
    \centering
    \includegraphics[width=\linewidth]{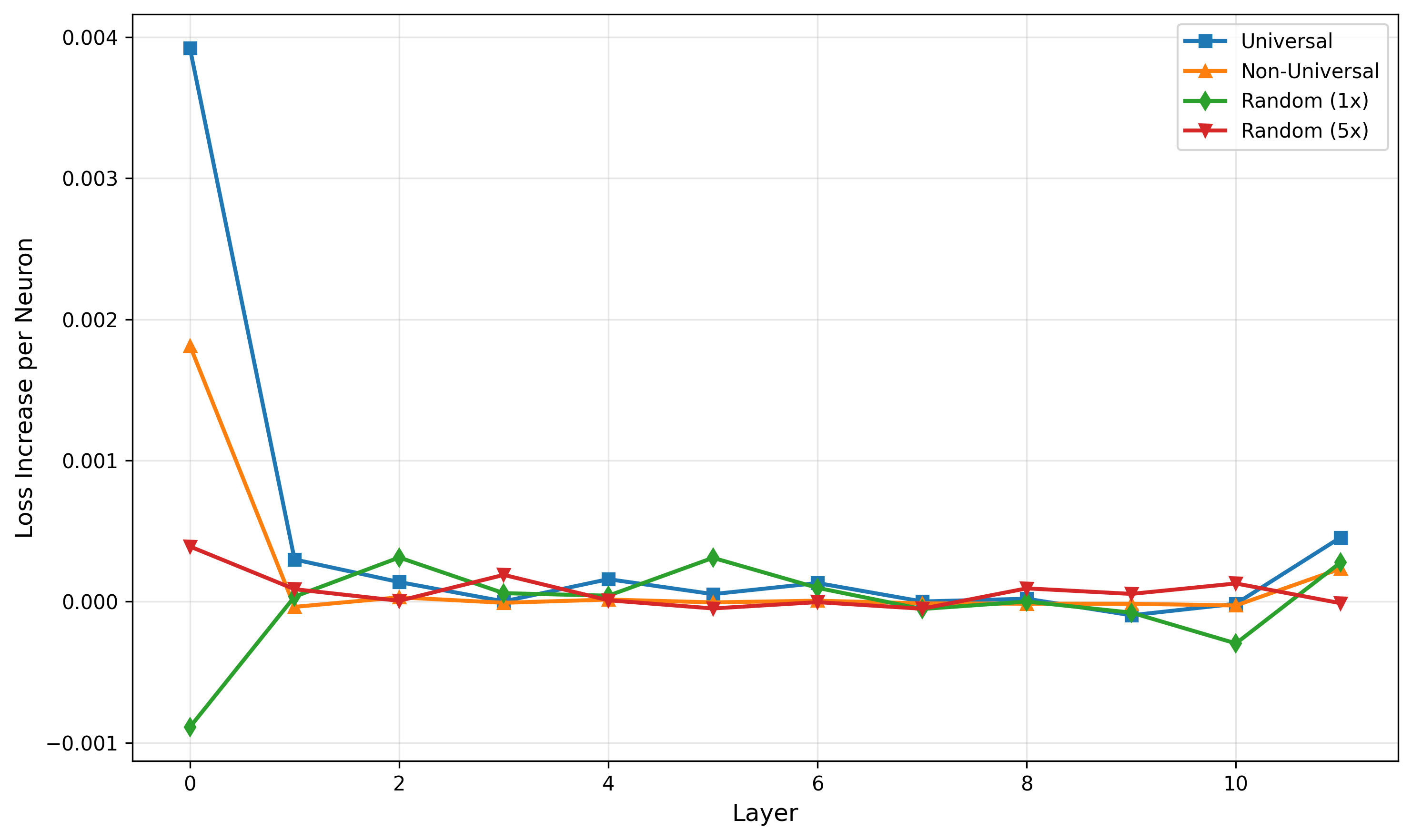}
    \caption{Layer-wise Ablation Efficiency (Change in loss per Neuron) on checkpoint 160k.}
  \end{minipage}
\end{figure}
\begin{figure}[htbp]
  \centering
  \begin{minipage}[t]{0.49\linewidth}
    \centering
    \includegraphics[width=\linewidth]{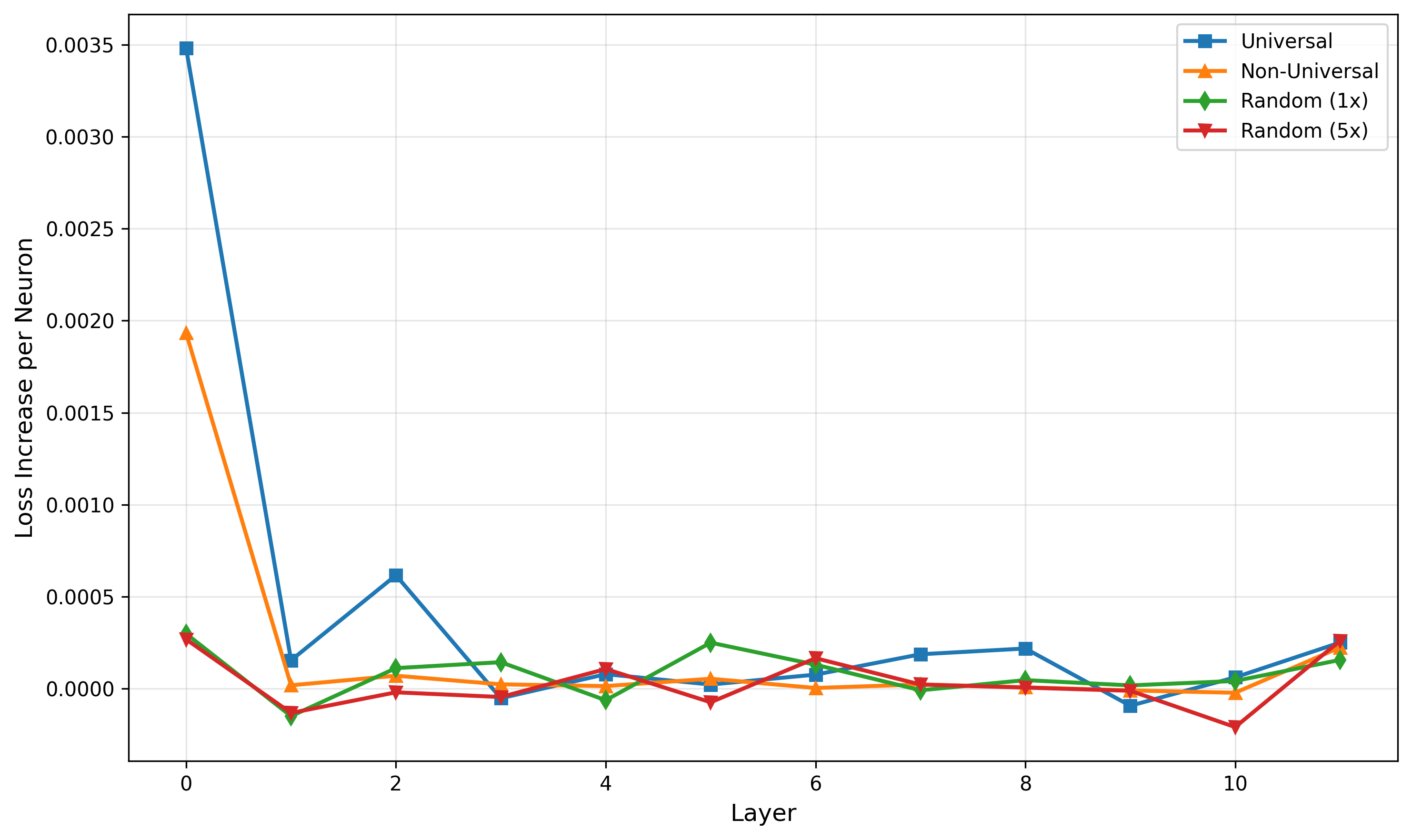}
    \caption{Layer-wise Ablation Efficiency (Change in loss per Neuron) on checkpoint 240k.}
  \end{minipage}%
  \hfill
  \begin{minipage}[t]{0.49\linewidth}
    \centering
    \includegraphics[width=\linewidth]{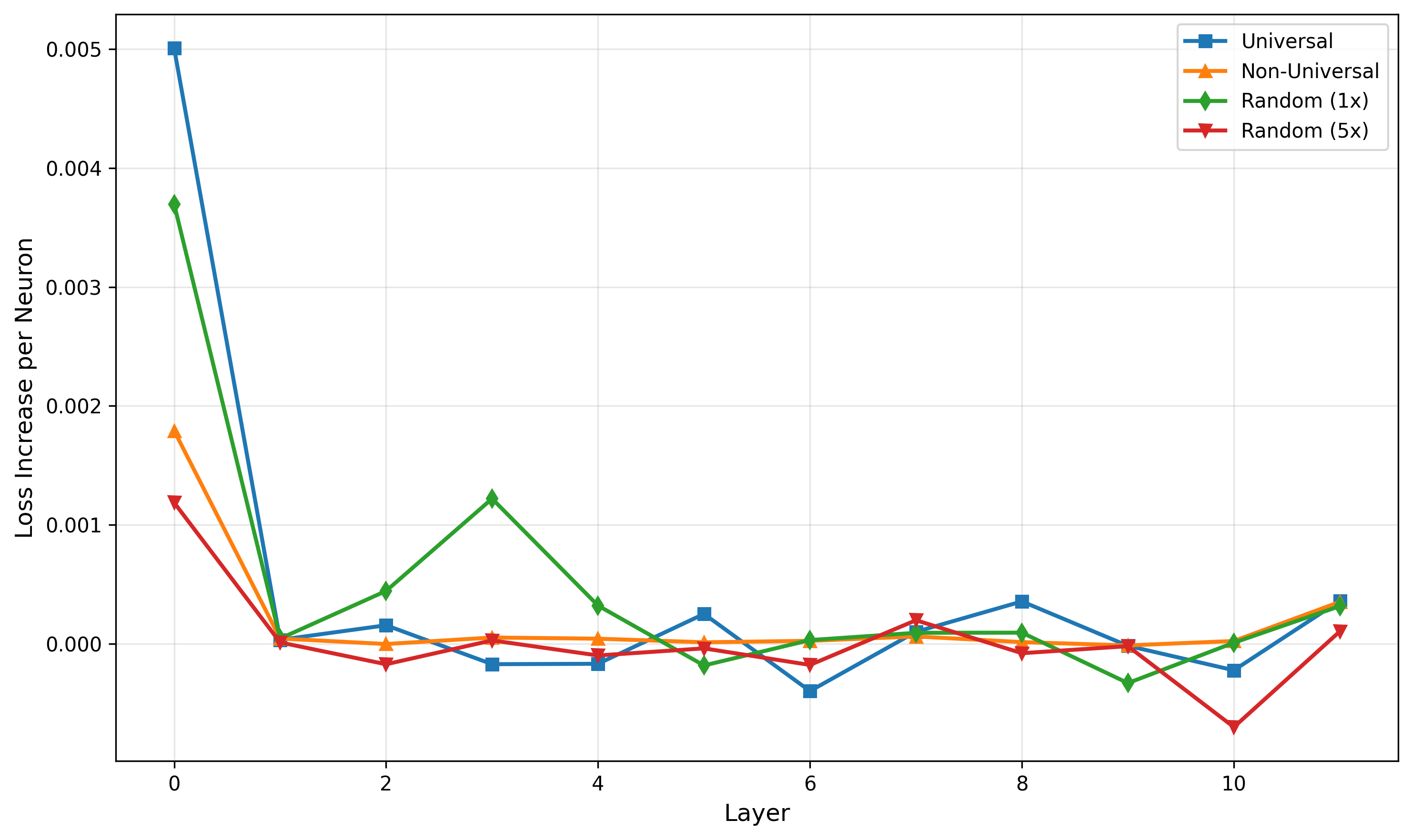}
    \caption{Layer-wise Ablation Efficiency (Change in loss per Neuron) on checkpoint 320k.}
  \end{minipage}
\end{figure}
\begin{figure}[htbp]
  \centering
    \includegraphics[width=0.49\linewidth]{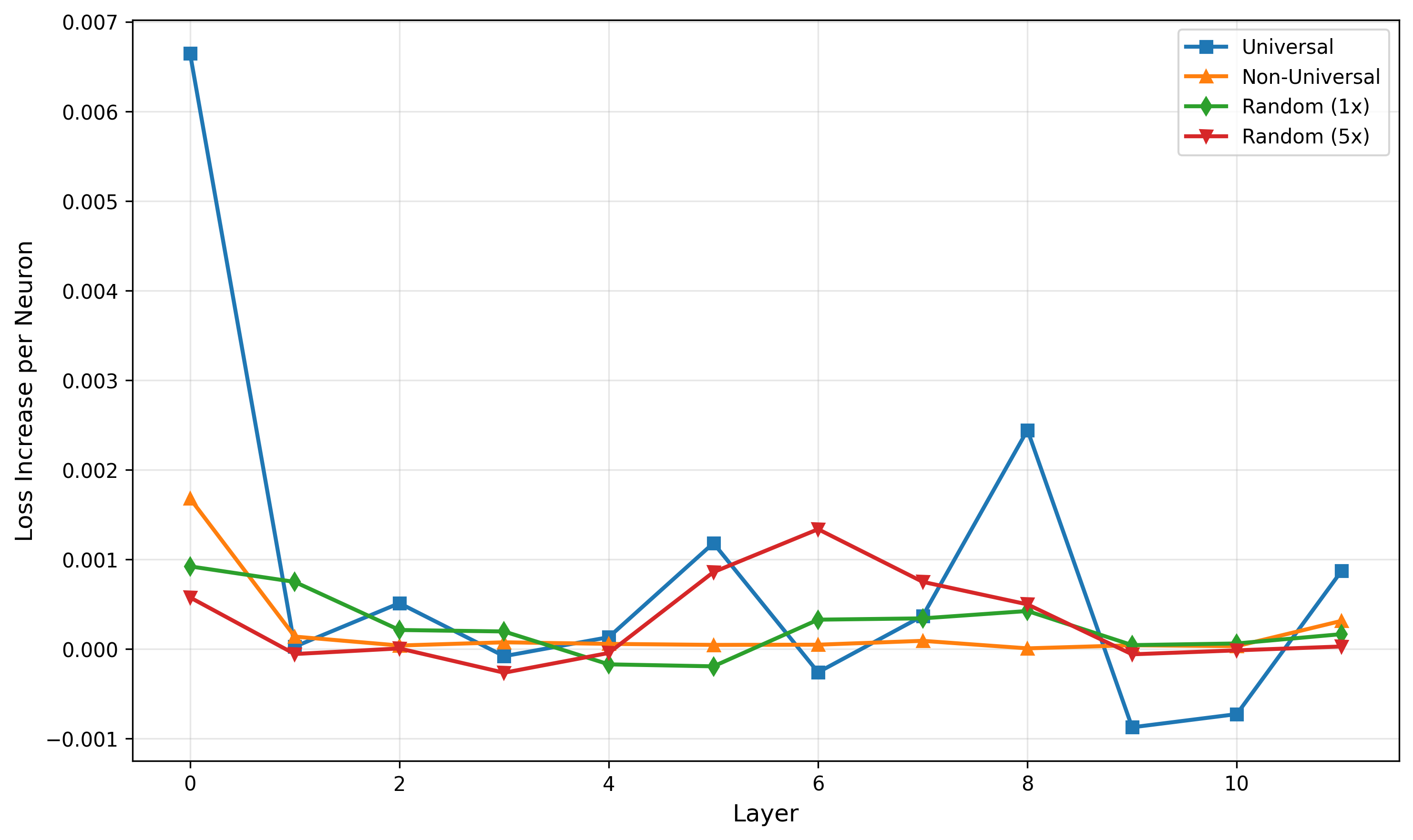}
    \caption{Layer-wise Ablation Efficiency (Change in loss per Neuron) on checkpoint 400k.}
\end{figure}

\end{document}